\documentclass[10pt,twocolumn,letterpaper]{article}

\usepackage{cvpr}              %

\usepackage{graphicx}
\usepackage{amsmath}
\usepackage{amssymb}
\usepackage{booktabs}

\usepackage{multirow}
\usepackage{subcaption}
\usepackage[table]{xcolor}

\newcommand{\myrowcolour}{\rowcolor[gray]{0.925}}
\newcommand{\vecb}[1]{\boldsymbol{#1}}

\usepackage[pagebackref,breaklinks,colorlinks]{hyperref}

\usepackage[capitalize]{cleveref}
\crefname{section}{Sec.}{Secs.}
\Crefname{section}{Section}{Sections}
\Crefname{table}{Table}{Tables}
\crefname{table}{Tab.}{Tabs.}

\def\cvprPaperID{} %
\def\confName{}
\def\confYear{}

\begin{document}

\title{PLIKS: A Pseudo-Linear Inverse Kinematic Solver for 3D Human Body Estimation}

\title{PLIKS: A Pseudo-Linear Inverse Kinematic Solver for 3D Human Body Estimation}

\author{%
{Karthik Shetty$^{1,2}$} ~ {Annette Birkhold$^2$}  ~ {Srikrishna Jaganathan$^{1,2}$}  {Norbert Strobel$^{2,3}$} \and ~  {Markus Kowarschik$^2$} ~  {Andreas Maier$^1$} ~ {Bernhard Egger$^1$} \\
\normalsize $^1$FAU~Erlangen-Nürnberg, Erlangen, Germany \quad
\normalsize $^2$Siemens~Healthineers~AG, Forchheim, Germany\\
\normalsize $^3$University of Applied Sciences W\"urzburg-Schweinfurt, Germany\\
{\tt\small karthik.shetty@fau.de}
}
\maketitle

\begin{abstract}
We introduce PLIKS (Pseudo-Linear Inverse Kinematic Solver) for reconstruction of a 3D mesh of the human body from a single 2D image. Current techniques directly regress the shape, pose, and translation of a parametric model from an input image through a non-linear mapping with minimal flexibility to any external influences. We approach the task as a model-in-the-loop optimization problem. PLIKS is built on a linearized formulation of the parametric SMPL model. Using PLIKS, we can analytically reconstruct the human model via 2D pixel-aligned vertices. This enables us with the flexibility to use accurate camera calibration information when available. PLIKS offers an easy way to introduce additional constraints such as shape and translation. We present quantitative evaluations which confirm that PLIKS achieves more accurate reconstruction with greater than $10\%$ improvement compared to other state-of-the-art methods with respect to the standard 3D human pose and shape benchmarks while also obtaining a reconstruction error improvement of $12.9~\mathrm{mm}$ on the newer AGORA dataset. 

\end{abstract}

\section{Introduction}

Estimating human surface meshes and poses from  single images is one of the core research directions in computer vision, allowing for multiple applications in computer graphics, robotics and augmented reality~\cite{LoPresti2016,Hogg1983ModelbasedVA}. Since humans have complex body articulations and the scene parameters are typically unknown, we are essentially dealing with an ill-posed problem that is difficult to solve in general.

\begin{figure}[!t]
\begin{center}
\includegraphics[width=0.95\linewidth,keepaspectratio]{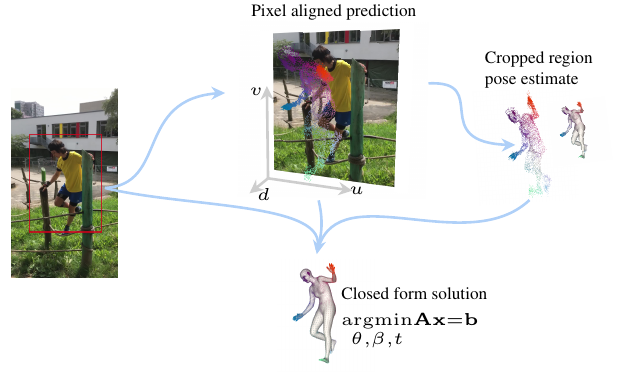}
\end{center}
   \caption{Network predicts a  pixel-aligned vertex map $(u,v,d)$ which is used to obtain an initial pose estimate. Then a closed-form solution is made use of to solve the Inverse kinematics between the 2D pixel-aligned vertex map $(u,v)$ and a pseudo-parametric model given the detected bounding-box camera intrinsic and initial pose estimate.   
   }
\label{fig-1}
\end{figure}

Thanks to models such as SMPL~\cite{smpl} and SMPL-X~\cite{SMPL-X:2019} additional constraints on body shape and pose became available. They made the problem somewhat more tractable. 
Most state-of-the-art methods~\cite{spin,eft,hmr,pare,romp} directly regress the shape and pose parameters from a given input image. These approaches rely completely on neural networks,
while making several assumptions about the image generation process. One typical assumption is the use of a  simplified camera model such as the weak perspective camera. In this scenario, the camera is assumed to be far away from the subject, which is generally realized by setting a large focal length constant for all  images. A weak perspective camera can be described based on  three parameters, two with respect to translation in the horizontal and vertical directions, and the third being scale. While these methods can estimate plausible shape and pose parameters, it can happen that the resulting meshes are either misaligned in the 2D image space or in the 3D object space. This is because the underlying optimization problem is often not constrained enough such that it is difficult for the underlying networks to optimize between the 2D re-projection loss and the 3D loss.

Some existing methods~\cite{wpc,spec,cliff} propose a workaround by tackling the problem using a hybrid approach involving learning-based and optimization-based techniques while incorporating a full perspective camera~\cite{wpc}. Optimization-based approaches are, however, prone to local minima, and they are computationally expensive. In~\cite{spec}, the authors propose to regress the SMPL parameters by conditioning on features from a CamCalib network meant to predict the camera parameters. Unfortunately, this camera prediction network needs a specialized dataset to train on, which is very hard to acquire in practice. It also prevents end-to-end learning.  

On the other hand, recent non-parametric or model-free approaches~\cite{meshgraph,i2l}  directly regress the mesh vertex coordinates based on their 2D projections, aligning well to the input image. However, by ignoring the effects of a perspective camera, even these methods suffer from the same limitations as the parametric models.

Motivated by the above observations, we present a novel approach, named PLIKS, for 3D human shape and pose estimation that incorporates the perspective camera while analytically solving for all the parameters of the parametric model. The pipeline of our approach comprises of two modules, namely the mesh regressor and PLIKS. The mesh regressor provides a mapping between an image and the 3D vertices of the SMPL model. Given a single image, any off-the-shelf Convolution Neural Network (CNN) can be used for feature extraction. The extracted features can then be used to obtain a mesh representation either by using 1D CNNs~\cite{i2l}, GraphCNNs~\cite{gcmr}, or even transformers~\cite{meshgraph}. This way, correspondences to the image space can be found and a relative depth estimate can be computed. From the image-aligned mesh prediction, we can roughly estimate the rotations with respect to a template mesh in canonical space with the application of Inverse Kinematics (IK), denoted in this work as the Approximate Rotation Estimator (ARE).
Finally, we reformulate the SMPL model as a linear system of equations, with which we can use the 2D pixel-aligned vertex maps and any known camera intrinsic parameters to fully estimate the model without the need for any additional optimization. As our approach is end-to-end differentiable and fits the model within the training loop, it is self-improving in nature. The proposed approach is benchmarked against various 3D human pose and shape datasets, and significantly outperforms other state-of-the-art approaches.

To summarize, the contribution of our paper is the following: (1) We bridge the gap between the 2D pixel-aligned vertex maps
and the parametric model by reformulating the SMPL model as a linear system of equations. Since the proposed approach is fully differentiable, we can perform end-to-end training. (2) We propose a 3D human body estimation framework that reconstructs the 3D
body without relying on weak-perspective assumptions. (3) We show that our approach can improve upon other state-of-the-art methods when evaluated across various 3D human pose and shape benchmarks.

\section{Related Work}
Recovering human pose and shape from monocular images has been extensively studied using both model-based~\cite{hmr,pare,romp} and model-free approaches~\cite{i2l,meshgraph,densepose}.  Model-based methods estimate the parameters of a parametric body model such as SMPL~\cite{smpl} based on a single input RGB image. The use of parametric body models makes it possible to enforce strong statistical priors of the human mesh. Model-based methods can be further split into optimization and regression techniques. Optimization-based approaches~\cite{bogo2016keep,mirror} make use of 2D keypoints estimated by a Deep Neural Network (DNN) which are iteratively fit with the SMPL model. These methods however are sensitive to initialization and are susceptible to local minima.
Regression-based techniques based on DNNs directly estimate the pose and shape parameters~\cite{hmr, guler2019holopose,omran2018neural,pavlakos2018learning,tung2017self,romp}. However, these approaches typically require a large amount of training data. SPIN~\cite{spin} provided a revolutionary architecture combining optimization-based techniques with regression-based methods. This allowed for much stronger supervision and improved performance on mesh accuracy. However, due to the difficulty in directly estimating the mapping from a single image to the shape and pose space, the mesh alignment with respect to the input image is often imperfect~\cite{hybrik,kama}. Model-free approaches directly regress the vertices based on intermediate representations, such as IUV maps~\cite{densepose,denserac,decomr}, 2D/3D heatmaps~\cite{kama,hybrik}, silhouette~\cite{straps,lasor}, and direct vertex regression~\cite{i2l,meshgraph,gcmr,l21end} where correspondence is established between the model and the input image. 

To overcome the issues of the learning-based and optimization-based approaches, hybrid techniques combining both approaches have been proposed~\cite{hybrik,llr,hkmr,kama}. In contrast to SPIN~\cite{spin}, here we have a closed-form solution. First, the learning-based approach localizes 3D human joint coordinates~\cite{rootnet,choi2022learning,Iqbal2020WeaklySupervised3H,p2m}, utilizing volumetric heatmaps or GraphCNNs for the target representation.
From the localized 3D joints, the swing rotations are then determined analytically, whereas the twist rotations, shape parameters, and root translation are either predicted by a network or iteratively optimized~\cite{hybrik,kama}.

Typical human reconstruction methods~\cite{hmr,spin} take a cropped input image while using 3D predictions projected onto the cropped image for 2D supervision.  This, however, ignores the effects of perspective warping. This happens when the cropped image is off-center resulting in inaccurate rotations~\cite{pcl}. To overcome this, some approaches~\cite{kama,wpc} perform iterative optimization ~\cite{bogo2016keep} after the initial DNN-based predictions. SPEC~\cite{spec} proposes to condition the image features from a camera calibration network. However, a few methods include perspective warping during the cropping process and add an implicit camera rotation as post-processing~\cite{mpiinf,pcl}. CLIFF~\cite{cliff} addresses the problem by incorporating the bounding box information into the cropped image.

\section{Methodology}
In this section, we present our network architecture which includes an analytical solver for inverse kinematics and is end-to-end trainable. As illustrated in Fig.~\ref{fig-arc}, our network consists of two parts, first, a mesh regressor and second, a Pseudo-Linear Inverse Kinematic Solver (PLIKS). In Sec.~\ref{meth-sec-001}, we go over the SMPL model along with its forward kinematics process, and in Sec.~\ref{meth-sec-002}, we explain the full pipeline for mesh reconstruction.

\subsection{Parametric Mesh Representation}\label{meth-sec-001}
We use the Skinned Multi-Person Linear (SMPL) model to parameterize the human body~\cite{smpl}. The SMPL model is a statistical parametric function $\mathcal{M}(\vecb{\beta}, \vecb{\theta}; \vecb{\Phi})$. The output of this function is a triangulated surface mesh with $N=6890$ vertices. 
The shape parameters $\vecb{\beta}$ are represented by a low dimensional principal component which maps the linear basis $\mathbf{B}$ from $\mathbb{R}^{|\vecb{\beta}|} \mapsto \mathbb{R}^{3 N}$, representing offsets to the average mesh  $\vecb{\Bar{x}}_m$ as $\vecb{x} {=} \vecb{\Bar{x}}_m {+} \vecb{\beta}\mathbf{B}$. The pose of the model is defined with the help of a kinematics chain involving a set of relative rotation vectors  $\vecb{\theta}=\left[\vecb{\theta}_{1}, \ldots, \vecb{\theta}_{K}\right] \in \mathbb{R}^{K \times 3}$ made up of $K = 24$ joints represented using axis-angle rotations. Additional model parameters summarized as $\mathbf{\Phi}$  are involved in the deformation process of the SMPL model. They are used as joint regressor $\mathcal{J}\in\mathbb{R}^{K\times 3N}$, blend weights $\mathcal{W}\in\mathbb{R}^{K\times N}$ and shape deformations conditioned on the body pose $B_{P}(\vecb{\theta})$. Starting from a mean template mesh, the desired body mesh is obtained by applying forward kinematics based on the relative rotations $\vecb{\theta}$ and shape deformations $\vecb{\beta}$. The 3D body joints can be obtained by a linear combination of the mesh vertices using any desired linear regressor $\mathcal{J}^{'}\left(\mathcal{M}(\vecb{\beta}, \vecb{\theta}; \vecb{\Phi})\right)$. 

\begin{figure}[!b]
\begin{center}
\includegraphics[width=0.4\linewidth,keepaspectratio]{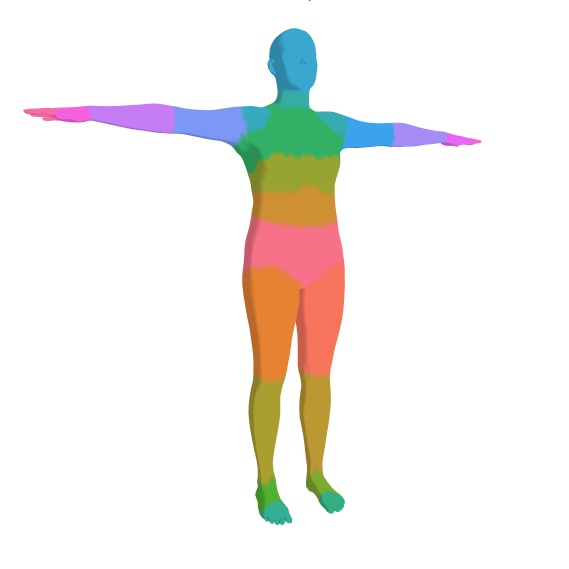}
\end{center}
  \caption{We split the SMPL model into 24 segments based on the blend weights. These segments help in determining the initial rotation estimate in the ARE pipeline.
   Here the segments are assumed to be rigid.}
\label{fig-2}
\end{figure}

\paragraph{Simplified SMPL}~\label{sec:ssmpl}
Although the SMPL model described above
is intrinsically linear, it cannot be defined as a linear system of equations to solve the IK due to the forward kinematics and pose-related shape deformations $B_{P}(\vecb{\theta})$. To deal with this problem, we make use of a simplified model, where we ignore the pose-related shape deformations $B_{P}(\vecb{\theta})$ in the formulation. We further split the SMPL mesh into $K{=}24$ segments $\mathbf{s}_{k \in (1,K)}$ corresponding to the influence of the 24 rotation vectors $\vecb{\theta}$ of the SMPL model as shown in Fig.~\ref{fig-2}. A vertex on the mesh with index $i$ belongs to a particular segment that has the maximum blend weight of all the joints, i.e., $\operatorname*{arg\,max}_{k \in (1,K)} \mathcal{W}(i)$. 
For ease of notation, we represent a set of vertices/indices corresponding to a segment $s_k$ with the superscript $k$.
Here, we also consider the individual segments to be a set of rigid 3D points which enables us to orient these segments from the template mesh  $\vecb{\Bar{x}}_m^k$ to a network predicted mesh $\mathbf{X}^k$ to obtain an initial rotation estimate using ARE explained in the following section.

\subsection{Pseudo-Linear Inverse Kinematic Solver}\label{meth-sec-002}
Solving Inverse Kinematics (IK) from 2D correspondences is a challenging task due to the inherent non-linearity that exists in the parameterized mesh generation process. We propose to solve IK from 2D pixel-aligned vertex inputs by building a linear system of equations of the form $\mathbf{A}\mathbf{x}{=}\mathbf{b}$. We add shape constraints to obtain the optimal world pose $\vecb{\hat{\theta}}$, shape $\vecb{\beta}$, and world translation $\vecb{t}$. Linear-Least-Squares is used to estimate the optimal parametric solution making the entire pipeline end-to-end differentiable. From the world pose $\vecb{\hat{\theta}}$, the relative rotations $\vecb{\theta}$ can be inferred by recursively solving the kinematic tree. For the linear system, we assume the rotations as a first-order Taylor approximation~\cite{taylor} with the individual segments of the predicted mesh model approximately oriented $\vecb{\Tilde{\theta}}$ along the optimal solution. Based on this assumption, using the rotations estimated from ARE should provide the exact solution in a single optimization step.

\begin{figure*}
\begin{center}
\centering
\includegraphics[width=\linewidth,keepaspectratio]{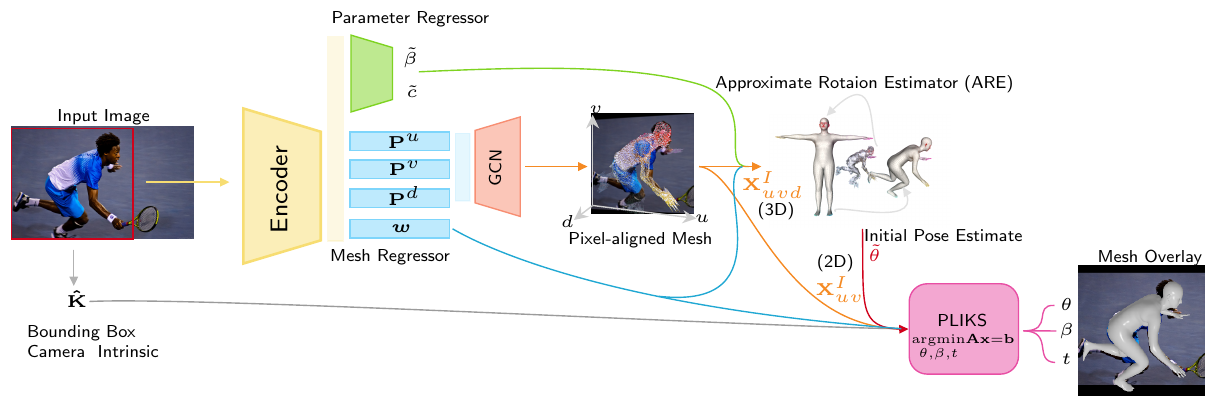}
\end{center}
  \caption{\textbf{Overview of the proposed framework:}  For a given RGB image input, the mesh regressor generates a mapping to the 3D mesh vertices aligned to the image plane. The ARE module provides rotation estimates in world space for each joint segment of the SMPL. The reconstructed body mesh is obtained via an analytical solution through the PLIKS process, fed with $\vecb{\Tilde{\theta}}$ comprising the segments' initial world rotations and the 2D mesh projections $\mathbf{X}^I_{uv}$ as inputs.}
\label{fig-arc}
\end{figure*}

\paragraph{Regression Network}
Assuming that some form of dense vertex correspondences exists that facilitates mapping between 3D vertices to pixels in the 2D image plane, we can then incorporate the IK solver into the network. As shown in Fig.~\ref{fig-arc}, our architecture is comprised of an encoder, a mesh regressor, and a parameter regressor. The encoder is based on HRNet-W32~\cite{hrnet}, the mesh regressor is based on MeshNet~\cite{i2l}, whereas the parameter regressor is a set of fully connected layers.  

The encoder acts as a feature extractor $\mathbf{F} \in \mathbb{R}^{C \times c^{'} \times c^{'}}$, with a channel dimension $C{=}480$, height and width $c^{'}{=}58$. It takes a cropped image $\mathbf{I} \in \mathbb{R}^{3 \times 224 \times 224}$ of a person as input. Similar to MeshNet~\cite{i2l}, we make use of $1\mathrm{D}$ convolutions to generate four feature vectors
\begin{equation*}
\mathbf{P} = \{\{\mathbf{P}^u,\mathbf{P}^v,\mathbf{P}^d\}\in \mathbb{R}^{N \times 58},\vecb{w}\in \mathbb{R}^{N}\},
\end{equation*}
for each mesh vertex. Here, $\mathbf{P}^u$ and $\mathbf{P}^v$ represent the features along the $u$ and $v$ axis, $\mathbf{P}^d$ represents the features along the root-normalized depth, and $\vecb{w}$ represents the weighting factor. They are obtained as follows 
\begin{equation}
\begin{aligned}
\mathbf{P}^u & = f_u^{1\mathrm{D}}(\mathrm{avg} ^u(\mathbf{F})),\\
\mathbf{P}^v & = f_v^{1\mathrm{D}}(\mathrm{avg}^v(\mathbf{F})),\\
\mathbf{P}^d & = f_d^{1\mathrm{D}}(\psi_d(\mathrm{avg}^{u,v}(\mathbf{F}))),\\ \vecb{w} & = \sigma(\mathrm{avg}^d(f_d^{1\mathrm{D}}(\psi_d(\mathrm{avg}^{u,v}(\mathbf{F}))))).
\end{aligned}\label{eq:feat}
\end{equation}
Here $f^{1\mathrm{D}}_i(.)$ and $\psi_i(.)$ represent a $1\mathrm{D}$ convolution along the $i\mathrm{{-}th}$ dimension, which converts the features $\mathbf{F}$ from $C {\times} c^{'} \to N {\times} c^{'}$ and  $C{\times} 1 \to (c^{'}{\times} C)^{T}$, respectively. The average function $\mathrm{avg}^i(.)$ averages the features along the $i\mathrm{{-}th}$ dimension. For the weighting factor $\vecb{w}$, we average across the channel dimension, followed by applying the sigmoid activation function $\sigma(.)$~\cite{sigmoid}. More details about the weighting function $\vecb{w}$ are explained in the following sections. 

We then concatenate and process $\Tilde{\mathbf{P}} {=} \{\mathbf{P}^u,\mathbf{P}^v,\mathbf{P}^d\}$ using a graph convolution network (GCN) to predict $uvd \in \mathbb{R}^{N \times 3}$. We use GCN rather than the heatmap-based approach from MeshNet~\cite{i2l} to avoid any truncation-based artifacts, which are prominent when partial images of humans are supplied as inputs. For the GCN we use the formulation from Kipf \etal ~\cite{gcn}, defined as $\mathbf{G} {=} \sigma( \Tilde{\mathbf{A}}\Tilde{\mathbf{P}}\mathbf{W})$, where  $\Tilde{\mathbf{A}} \in \mathbb{R}^{N \times N}$ denotes the graph adjacency matrix, $\mathbf{W} \in \mathbb{R}^{c^{'}{\times}l}$ denotes the trainable weights with $l$ as the output channel dimension, and $\sigma$ denotes the ReLU~\cite{relu} activation function. We make use of $3$ GCNs in series, with channel sizes $64$, $32$, and $3$ respectively. The final output $\mathbf{X}^I_{uvd} \in \mathbb{R}^{N \times 3}$ acts as the vertex correspondence in the image coordinate system.

The parameter regressor is a set of fully connected layers to obtain an approximate shape $\vecb{\Tilde{\beta}}$ and a weak perspective camera $\vecb{\Tilde{c}} \in \mathbb{R}^3$, which are used to determine the approximate world rotations later. We make use of the estimated depth $\vecb{\Tilde{c}}_d$ from the camera prediction to obtain a mesh in the world coordinate system $\mathbf{X}$ as follows,
\begin{equation}
    \mathbf{X} = (\vecb{K}^{-1}\mathbf{X}^I_{uv1})\cdot(\mathbf{X}^I_{d}+\vecb{\Tilde{c}}_d), 
\end{equation}
where $\vecb{K} \in \mathbb{R}^{3\times 3}$ is an intrinsic matrix with fixed  focal length of $1000$.

\paragraph{Approximate Rotation Estimator (ARE)}
Given two sets of corresponding points in $3\mathrm{D}$ space, it is possible to obtain an optimal rotation as a closed-form solution using the Kabsch solver~\cite{kabsch}. For a given segment from network mesh prediction $\mathbf{X}^k$ we make use of the Kabsch solver to determine the rotation that a same segment from the template mesh needs to go through from its rest pose. As the mesh prediction $\mathbf{X}^k$ can represent a wide range of human shapes, we make use of the shape predictions $\vecb{\Tilde{\beta}}$ on the mean shape as $\vecb{x}^k {=} \vecb{\Bar{x}}_m^k {+} (\vecb{\Tilde{\beta}}\mathbf{B})^k$ . The pose solver minimizes the squared distances between a set of $3\mathrm{D}$ correspondences to obtain an optimal pose as follows,
\begin{equation}\label{svd_rot}
    (\vecb{U \Sigma V^{T}})_k {=} \vecb{C}_k {=}\sum_{\left(\mathbf{X}^k, \mathbf{x}^k\right)}\left(\mathbf{X}^k-\Bar{\mathbf{X}}^k\right)\left(\mathbf{x}^k-\Bar{\mathbf{x}}^k\right)^{T}\vecb{w}^k\mathcal{W}^k.
\end{equation}
Here $\vecb{C}_k$ is the covariance matrix between the correspondences. In this context, $\Bar{\mathbf{x}}^k$ and $\Bar{\mathbf{X}}^k$ represent the means over the associated point sets. The world rotation for the $k$-th segment $\vecb{\Tilde{\theta}}_k {=} (\vecb{VU}^T)_k$ is obtained by applying singular value decomposition (SVD) over the covariance matrix. Since the SVD is differentiable, gradients can be back-propagated during the training process. Due to the blend skinning $\mathcal{W}$ process, a vertex may have rotation influence from on one or more joints. To tackle this, we multiply the squared distances in Eq.~\eqref{svd_rot} with a weight term $\vecb{w}^k$ from Eq.~\eqref{eq:feat} and the maximum blend influence $\mathcal{W}^k$. The weight is learned in an unsupervised manner to offset any influence the blending process contributes.

The obtained rotations are in the world space, whereas the SMPL model expects relative rotations in its axis space. The rotation around the pelvis $\vecb{\Tilde{\theta}}_{k=1}$ corresponds to the global root rotation. The relative rotations for the other joints can be recursively based on the parent rotation following a pre-defined kinematic tree for the SMPL model. To this end, let $\mathbf{\Tilde{R}}_k = \mathcal{R} (\vecb{\Tilde{\theta}}_{k}) \in \mathbb{S}\mathbb{O}(3)$ represent the estimated world rotation determined for segment $k$. Then the relative rotation for segment $k$ is
\begin{equation}\label{eq_ik}
 \vecb{\hat{\theta}}_{k} =   \mathbf{\hat{R}}_k =  {\mathbf{\Tilde{R}}_{p(k)}^{{-}1}}\mathbf{\Tilde{R}}_k,
\end{equation} 
where $p(k)$ represents the parent joint of $k$ and $\vecb{\hat{\theta}}_{k}$ the estimated relative pose.

\paragraph{Pseudo-Linear Inverse Kinematic Solver (PLIKS)}
An approximate SMPL model projected onto the image plane, with no pose-related blend shapes can be represented as
\begin{equation}\label{eq_smpl_full}
    \vecb{i}^k =\mathbf{\hat{K}}  \sum_{j=1}^{K} \mathcal{W}^k_j\left( \mathbf{{\Delta R}}_k\mathbf{\Tilde{R}}_k(\vecb{\Bar{x}}_m^k +  \vecb{\beta}\mathbf{B}^k) + \vecb{t}_k\right),
\end{equation}

\begin{table*}[ht]
\centering
\footnotesize
\begin{tabular}{l|llllllll}
\multirow{2}{*}{Method} & \multicolumn{3}{c}{3DPW (14)}                          & \multicolumn{2}{c}{Human3.6M (14)}     & \multicolumn{3}{c}{MPI-INF-3DHP (17)} \\ \cline{2-9} 
 &
  \multicolumn{1}{c}{PA-MPJPE$\downarrow$} &
  \multicolumn{1}{c}{MPJPE$\downarrow$} &
  \multicolumn{1}{c|}{PVE$\downarrow$} &
  \multicolumn{1}{c}{PA-MPJPE$\downarrow$} &
  \multicolumn{1}{c|}{MPJPE$\downarrow$} &
  \multicolumn{1}{c}{PCK$\uparrow$} &
  \multicolumn{1}{c}{AUC$\uparrow$} &
  \multicolumn{1}{c}{MPJPE$\downarrow$} \\ \hline
HMR~\cite{hmr}                     & 81.3 & 130.0 & \multicolumn{1}{l|}{-}             & 56.8 & \multicolumn{1}{l|}{88.0} & -              & -   & -              \\\myrowcolour
SPIN~\cite{spin}                    & 59.2 & 96.9  & \multicolumn{1}{l|}{116.4}         & 41.1 & \multicolumn{1}{l|}{-}    & 76.4          & 37.1     & 105.2      \\
I2L$_{\mathrm{+}}$~\cite{i2l}        & 58.6 & 93.2  & \multicolumn{1}{l|}{-}             & 41.7 & \multicolumn{1}{l|}{55.7} & -              & -    & -            \\\myrowcolour
EFT$^{\dagger}$~\cite{eft}                    & 51.6 & -     & \multicolumn{1}{l|}{-}             & 44.0 & \multicolumn{1}{l|}{-}    & -           & -       & -         \\
ROMP$^{\dagger}$~\cite{romp}                   & 47.3 & 76.7  & \multicolumn{1}{l|}{93.4}          & -    & \multicolumn{1}{l|}{-}    & - & -               & 95.1               \\\myrowcolour

PARE$^{\dagger}$~\cite{pare}                   & 46.5 & 74.5  & \multicolumn{1}{l|}{88.6}          & -    & \multicolumn{1}{l|}{-}    & -              & -    & -            \\
Mesh Graphormer$^{\dagger}_\mathrm{+}$~\cite{meshgraph}         & 45.6 & 74.7  & \multicolumn{1}{l|}{87.7}          & 34.5 & \multicolumn{1}{l|}{51.2} & -              & -   & -             \\\myrowcolour
HybrIK$^{\dagger}$~\cite{hybrik}                 & 45.3 & 74.1  & \multicolumn{1}{l|}{86.5} & 33.6 & \multicolumn{1}{l|}{55.4} & 87.5     & 46.9         & 93.9         \\ 
CLIFF$^{\dagger}$~\cite{cliff}                 & 43.0 & 69.0  & \multicolumn{1}{l|}{81.2} & \textbf{32.7} & \multicolumn{1}{l|}{47.1} & -     & -         & -            \\ \hline

PLIKS$^{\dagger}$&
  42.8 &
  66.9 &
  \multicolumn{1}{l|}{82.6} &
  34.7 &
  \multicolumn{1}{l|}{49.3} &
  91.8 & 52.3 &
  72.3 \\ \myrowcolour 
  
PLIKS$^{\ddag}$&
  {40.1} &
  {63.5} &
  \multicolumn{1}{l|}{{76.7}} &
  34.9 &
  \multicolumn{1}{l|}{\textbf{46.8}} &
  {93.1} & {53.0} &
  {69.8} \\ 
  
   PLIKS$^{\ddag}$(HR48)&
   \textbf{38.5} &
   \textbf{60.5} &
   \multicolumn{1}{l|}{\textbf{73.3}} &
   34.5 &
   \multicolumn{1}{l|}{{47.0}} &
   \textbf{93.9} & \textbf{54.1} &
   \textbf{67.6} \\ 
  \hline

 \end{tabular}\caption{\label{tab:main} Benchmark of state-of-the-art models on 3DPW, Human3.6M, and MPI-INF-3DHP datasets. All units are in mm. Here $+$ represents non-parametric methods. $\dagger$ means that the network was additionally trained with 3DPW. $\ddag$ means that the network was additionally trained with 3DPW and AGORA.
}
\end{table*}

Here, the superscript $k$ refers to the $k$-th's subset of vertices defined in Sec.~\ref{sec:ssmpl}. The matrix  $\mathbf{\hat{K}} \in \mathbb{R}^{3\times 4}$ represents a perspective projection matrix taking into account the affine transformation (only crop and resize) for the image fed into the network as

\begin{equation*}
\mathbf{\hat{K}} = \left[\begin{array}{cccc}
f_x & 0 & p_x & 0\\
0 & f_y & p_y & 0\\
0 & 0 & 1 & 0
\end{array}\right].
\end{equation*}

 Further, $\mathbf{\Tilde{R}}_k$ represents the approximate world rotation obtained from the previous step,  $\mathbf{\Delta R}_k$ is defined as the additional rotation required to get an optimal solution, $\vecb{t}^k$ represents the joint translation, and $\mathcal{W}$ are the blend weights. By combining all segments, we get $\vecb{i} = \mathbf{X}^I_{uv1} \in \mathbb{R}^3$. This represents the correspondence between mesh vertices, defined in homogeneous coordinates, and the pixels in the image plane . 
 
 Since the additional rotation required is considered to be small we linearize the rotation matrix which needs to be determined based on Taylor expansion with angles $\alpha$, $\beta$, $\gamma$ along the $x$, $y$ and $z$ axes, respectively,  as follows 
 \begin{equation*}
     \mathbf{{\Delta R}}_k = \left[\begin{array}{cccc}
1 & -\gamma_k &\beta_k & 0 \\
\gamma_k & 1 & -\alpha_k & 0 \\
-\beta_k & \alpha_k & 1 & 0 \\
0 & 0 & 0 & 1 \\
\end{array}\right].
 \end{equation*}
 
 Further, we simplify the projected SMPL model in Eq.~\eqref{eq_smpl_full} by making some additional assumptions using the definitions introduced in Eq.~\eqref{eq_smpl_linear}. For the $\vecb{x}_r^k$ term we assume that the majority of the rotation is significantly effected by the rotation with respect to the primary segment, i.e., we ignore the impact of neighbouring rotations as they are usually minuscule. This assumption is also made for the term ${\mathbf{B}_r^k}$ by assuming that for small rotations $\mathbf{{\Delta R}}_k\mathbf{\Tilde{R}}_k\vecb{\beta} \mathbf{B}^k \approx \mathbf{\Tilde{R}}_k\vecb{\beta} \mathbf{B}^k$. 
\begin{equation}\label{eq_smpl_linear}
    \vecb{i}^k{=}\mathbf{\hat{K}} \Bigg( \mathbf{{\Delta R}}_k \underbrace{\sum_{j=1}^{K}\mathcal{W}_j^k \mathbf{\Tilde{R}}_k\vecb{\Bar{x}}_m^k}_{\vecb{x}_r^k}   + \vecb{\beta} \underbrace{\sum_{j=1}^{K}\mathcal{W}_j^k \mathbf{\Tilde{R}}_k \mathbf{B}^k}_{\mathbf{B}_r^k}  + \vecb{t}\underbrace{_k\sum_{j=1}^{K}\mathcal{W}_j^k}_{\mathbf{W}_r^k} \Bigg).
\end{equation}

The equation is now linear, with $154$ unknown parameters corresponding to $\mathbf{{\Delta R}}_k$, $\vecb{\beta}$ and $\vecb{t}_k$. If the focal lengths $f_x,f_y$ and principal points $p_x,p_y$ are not known, we assume a fixed focal length of $1000~\mathrm{mm}$ and the image center as the principal point. Note that the fixed values here apply to the entire image, i.e., not to the cropped and resized input fed to the network.

Using Direct Linear Transform (DLT)~\cite{dlt} we can rewrite Eq.~\eqref{eq_smpl_linear} in the form $\mathbf{A}\mathbf{x}=\mathbf{b}$. The optimal parameters can be obtained by minimizing the analytical error using linear least square defined by $\mathbf{x} = \mathbf{A}^{+}\mathbf{b}$, where $\mathbf{A}^{+}$ represents the pseudo-inverse of $\mathbf{A}$. As the pseudo-inverse of a tall matrix is differentiable~\cite{pinv}, gradients can be back-propagated during the training process. One of the major drawbacks when using DLT is that it minimizes the analytical loss rather than the geometric loss. One option to overcome this is to use Iterative Reweighted Least Squares (IRLS)~\cite{irls}, which robustly minimizes the objective function in an iterative manner by reweighing the geometric loss. However, we make use of the network predicted weighting $\vecb{w}^k$ to weight the correspondences~\cite{lookma,dfm}. To further enforce the predicted shape to be close to the mean shape, we add an additional constraint such that $||\vecb{\beta}||_2 \approx 0$ with a  regularizing weight $\omega_\beta$ as
\begin{equation}\label{eq_argmin}
    \underset{ \mathbf{{\Delta R}}_k,\vecb{\beta},\vecb{t}_k}{\mathrm{argmin}}|| \vecb{w}^k\Big(\vecb{i}^k - \mathbf{\hat{K}} ( \mathbf{{\Delta R}}_k {\vecb{x}_r^k}   + \vecb{\beta} {\mathbf{B}_r^k}  +  \vecb{t}_k\mathbf{W}_r^k)\Big)||_2 {+} \omega_\beta||\vecb{\beta}||_2.
\end{equation}

To get the final relative pose $\vecb{{\theta}}_k$ for each joint $k$, we apply Eq.~\eqref{eq_ik} on the obtained world rotations $\mathbf{{\Delta R}}_k \mathbf{\Tilde{R}}_k$. The global translation for the camera system $\mathbf{\hat{K}}$ is obtained from the root joint as $\vecb{t}_{1}{-}(\vecb{j}_{1}{-}\mathbf{{\Delta R}}_1 \mathbf{\Tilde{R}}_1\vecb{j}_{1})$, where $\vecb{j}_{1}$ is the root joint for a rest pose mesh with shape coefficient $\vecb{\beta}$.

\section{Experiments}
Following previous works, the base PLIKS model is trained on a combination of 3D datasets (Human3.6M~\cite{h36m} and MPI-INF-3DHP~\cite{mpiinf}), and 2D dataset (COCO~\cite{coco}) with pseudo-GT labels obtained from EFT~\cite{eft}. We evaluate on Human3.6M~\cite{h36m}, 3DPW~\cite{3dpw}, MPI-INF-3DHP~\cite{mpiinf}, AGORA~\cite{agora}, MuPoTs-3D~\cite{mupo}, and 3DOH50K~\cite{3doh}. During the evaluation, we highlight our results for networks trained with any additional datasets.

\begin{table}[]
\footnotesize
\centering
\begin{tabular}{l|cccc}
      & \multicolumn{2}{c}{3DPW}          & \multicolumn{2}{c}{AGORA} \\ \cline{2-5} 
      & MPJPE$\downarrow$ & \multicolumn{1}{c|}{PVE$\downarrow$}  & MPJPE$\downarrow$        & PVE$\downarrow$        \\ \hline
ARE   & 23.0  & \multicolumn{1}{c|}{24.9} & 52.0         & 55.3       \\
PLIKS & 1.3   & \multicolumn{1}{c|}{1.5}  & 8.4(1.1)          & 10.3(1.7)       \\\hline
\end{tabular}\caption{\label{tab:pea}Ground truth errors to validate the closed-form solution of PLIKS. Results in brackets represent running the PLIKS module twice. All units are in mm.}
\end{table}

\subsection{PLIKS Error Analysis}
 We measure the effective reconstruction capability of ARE and PLIKS with ground truth conditions for the 3DPW~\cite{3dpw} test set and the AGORA~\cite{agora} validation set. For the ARE module experiment, we provide the GT SMPL mesh vertex projections and the GT shape parameters. Additionally, we set the root depth as $7~\mathrm{m}$ for all the images as the SMPL model in its template pose can be well represented inside a $224\times224$ image when assuming weak perspective settings with a focal length of $1~\mathrm{m}$. For the PLIKS module experiment, we only provide the GT SMPL mesh vertex projections and a root depth of $7~\mathrm{m}$ to the ARE module along with the GT bounding-box camera intrinsics as inputs. We also set the weights for the shape regularizer $\omega_\beta {=} 0$.  We measure the Mean Per Joint Position Error (MPJPE), and the Per Vertex Error (PVE). Following previous works~\cite{spin,hmr,pare}, we use the LSP joint regressor~\cite{spin} to determine the 14 joints which are regressed from the body mesh. We report the results in Table~\ref{tab:pea} showing that the assumptions made in Eq.~\eqref{eq_smpl_linear} are reasonable providing a stable and accurate fit. We observe a larger error on the AGORA dataset with the PLIKS module, due to larger perspective warping effects, which results in incorrect global rotation estimation from the ARE module. As PLIKS uses a linearized rotation matrix, running the PLIKS module twice, that is iteratively optimizing the rotation estimates provides a more accurate fit. Though ARE is not an accurate IK solver, we can safely conclude about the drawbacks of using a weak perspective camera model.

\begin{table}[]
\footnotesize
\centering
\begin{tabular}{lllll}

                             & MRPE           & MRPE$_x$       & MRPE$_y$       & MRPE$_z$        \\ \hline
\multicolumn{1}{l|}{Baseline~\cite{wpc,lcr}} & 267.8          & 27.5          & 28.3          & 261.9          \\\myrowcolour
\multicolumn{1}{l|}{RootNet~\cite{rootnet}}  & 120.0 & 23.3          & 23.0          & 108.1 \\
\multicolumn{1}{l|}{PLIKS}     & 135.5          & 18.0 & \textbf{14.1} & 128.9          \\\myrowcolour
\multicolumn{1}{l|}{PLIKS$^\ddag$}     & \textbf{96.1}          & \textbf{16.1} & 14.7 & \textbf{88.5}          \\ \hline

\end{tabular}\caption{\label{tab:root}Evaluation on Human3.6M dataset, with respect to the MRPE and MRPE on $x$, $y$ and $z$ axis. $^\ddag$ means the network was additionally trained with 3DPW~\cite{3dpw} and AGORA~\cite{agora}.}
\end{table}

\begin{table}[]
\centering
\footnotesize
\begin{tabular}{l|cc|c}
      & \multicolumn{2}{c|}{Matched People}       & All People    \\ \cline{2-4} 
      & PCK$_{abs}$$\uparrow$                     & PCK$_{root}$$\uparrow$                      & PCK$_{abs}$$\uparrow$                           \\ \hline
Rootnet~\cite{rootnet}   & {31.8} & {31.0} & {31.5}       \\\myrowcolour
VirtualPose~\cite{virtualpose}    & \textbf{47.0}            & 53.5                      & 44.0                           \\
PLIKS$^\ddag$ & 44.8                     & \textbf{55.7}             & \textbf{44.2}                  \\ \hline
\end{tabular}\caption{\label{tab:root2}Absolute PCK evaluation on the MuPoTs-3D~\cite{mupo} dataset.}
\end{table}

\subsection{Comparison with the State-of-the-art}
We compare our method with previous human mesh reconstruction approaches based on Human3.6M, MPI-INF-3DHP, and 3DPW datasets. To remain consistent with previous approaches, we use the LSP regressor~\cite{spin} to obtain the 14 joints for Human3.6M and 3DPW datasets, while we use 17 joints with the Human3.6M regressor~\cite{spin} for the MPI-INF-3DHP dataset. We measure Procrustes Aligned Mean Per Joint Position Error (PA-MPJPE), Percentage of Correct Keypoints (PCK), and Area
Under Curve (AUC) on the 3D pose results.

\begin{figure*}[ht]
	\centering
	\begin{subfigure}[b]{0.195\textwidth}
	\centering
	    \includegraphics[width=\textwidth,keepaspectratio]{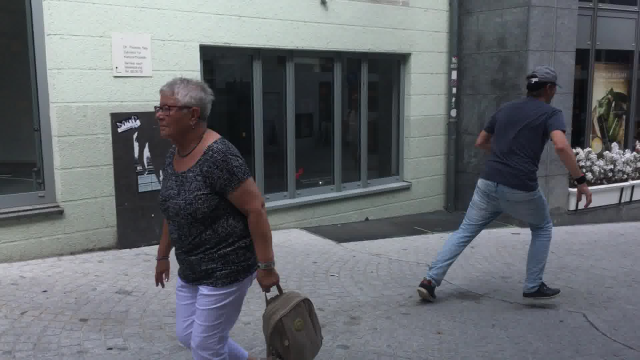}
         \includegraphics[width=\textwidth,keepaspectratio]{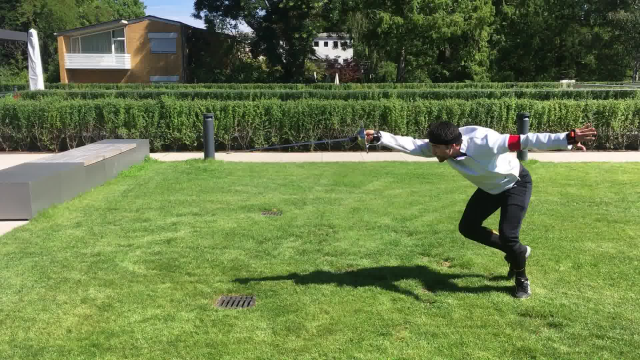}
         \includegraphics[width=\textwidth,keepaspectratio]{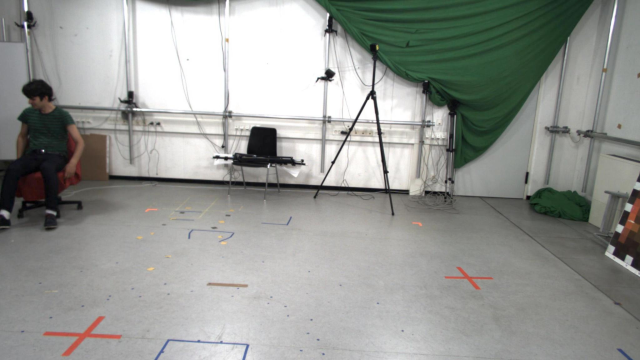}
         \includegraphics[width=\textwidth,keepaspectratio]{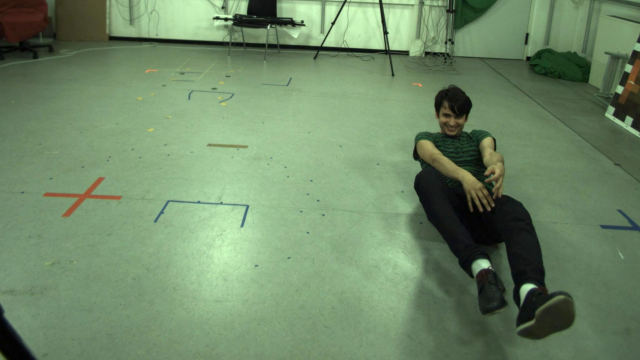}

    \caption*{Input Image}
    \end{subfigure}
    \begin{subfigure}[b]{0.195\textwidth}
	\centering
	     \includegraphics[width=\textwidth,keepaspectratio]{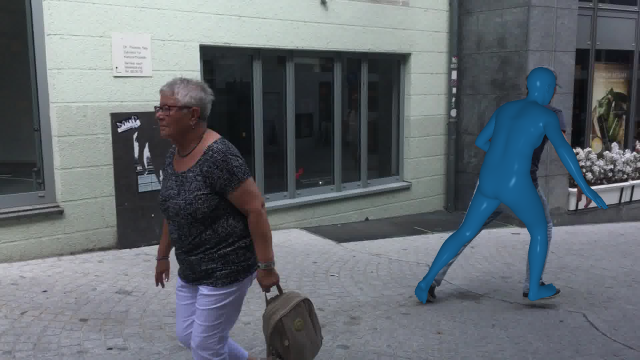}
         \includegraphics[width=\textwidth,keepaspectratio]{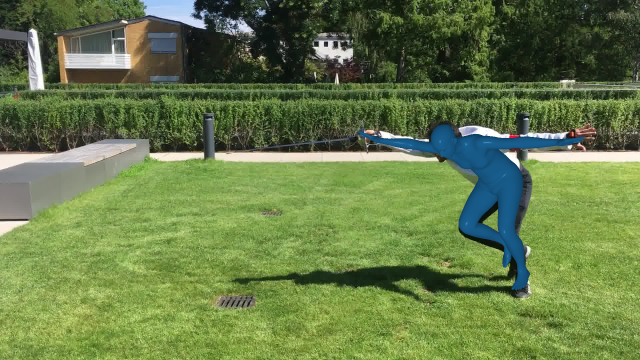}
         \includegraphics[width=\textwidth,keepaspectratio]{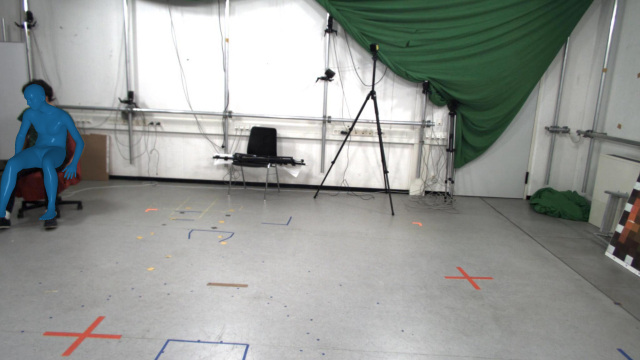}
         \includegraphics[width=\textwidth,keepaspectratio]{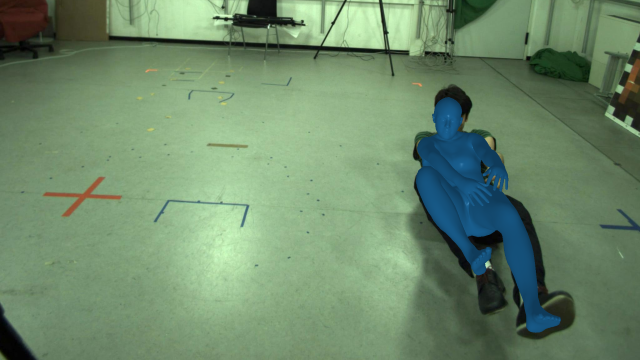}
    \caption*{PLIKS}
    \end{subfigure}
        \begin{subfigure}[b]{0.195\textwidth}
	\centering
	     \includegraphics[width=\textwidth,keepaspectratio]{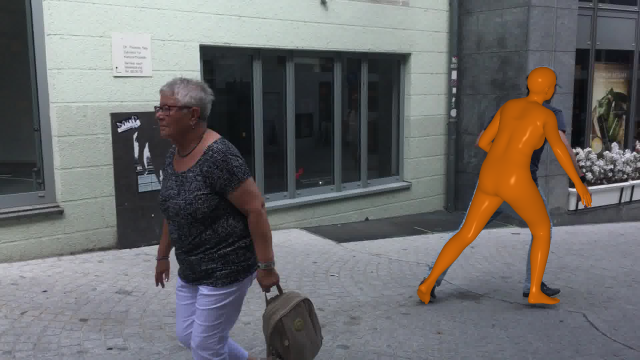}
         \includegraphics[width=\textwidth,keepaspectratio]{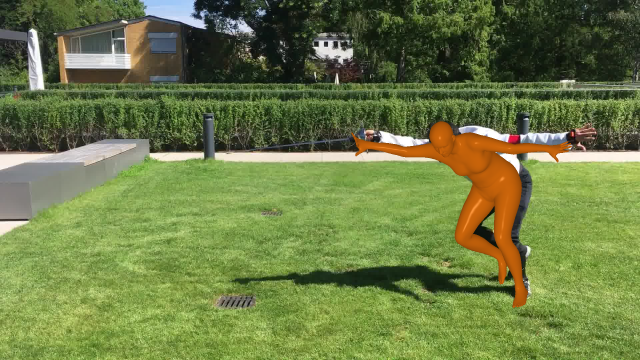}
         \includegraphics[width=\textwidth,keepaspectratio]{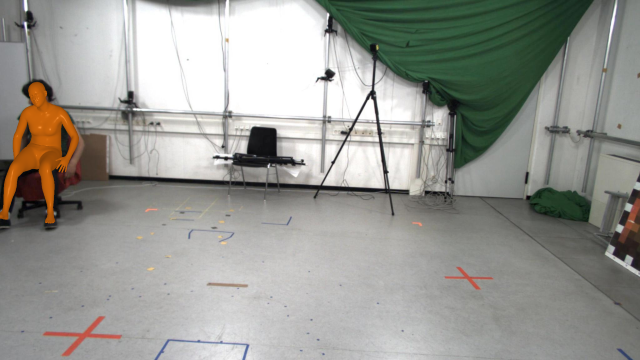}
         \includegraphics[width=\textwidth,keepaspectratio]{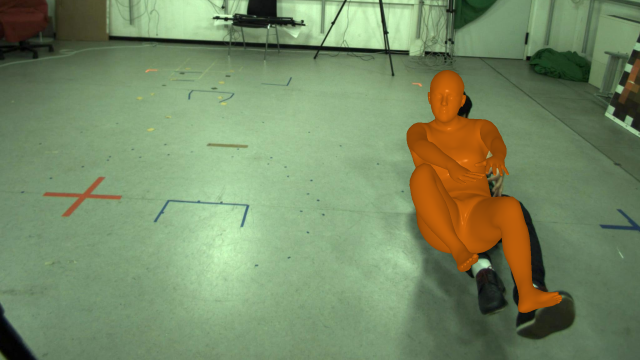}
    \caption*{HybrIK~\cite{hybrik}}
    \end{subfigure}
	\begin{subfigure}[b]{0.195\textwidth}
	\centering
	     \includegraphics[width=\textwidth,keepaspectratio]{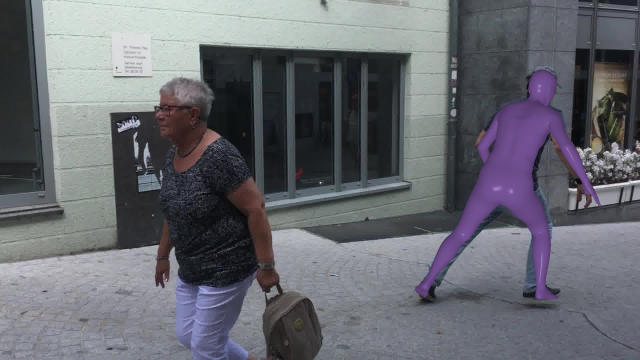}
         \includegraphics[width=\textwidth,keepaspectratio]{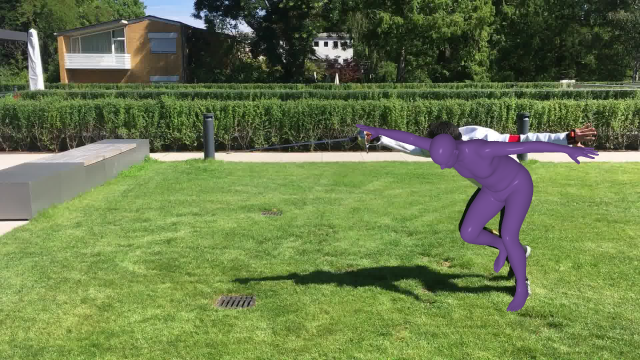}
         \includegraphics[width=\textwidth,keepaspectratio]{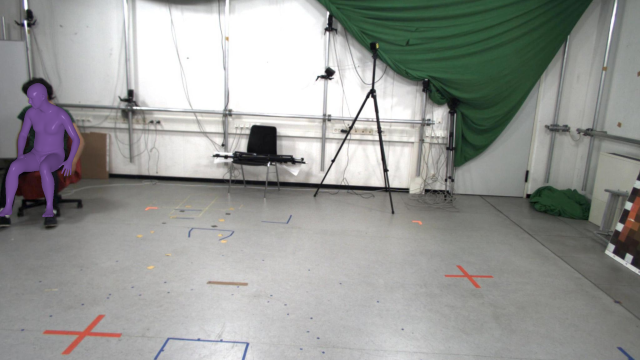}
         \includegraphics[width=\textwidth,keepaspectratio]{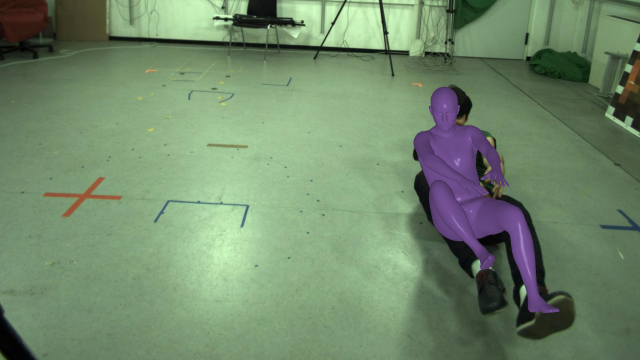}
    \caption*{CLIFF~\cite{cliff}}
    \end{subfigure}
    \begin{subfigure}[b]{0.1096875\textwidth}
	\centering
	     \includegraphics[width=\textwidth,keepaspectratio]{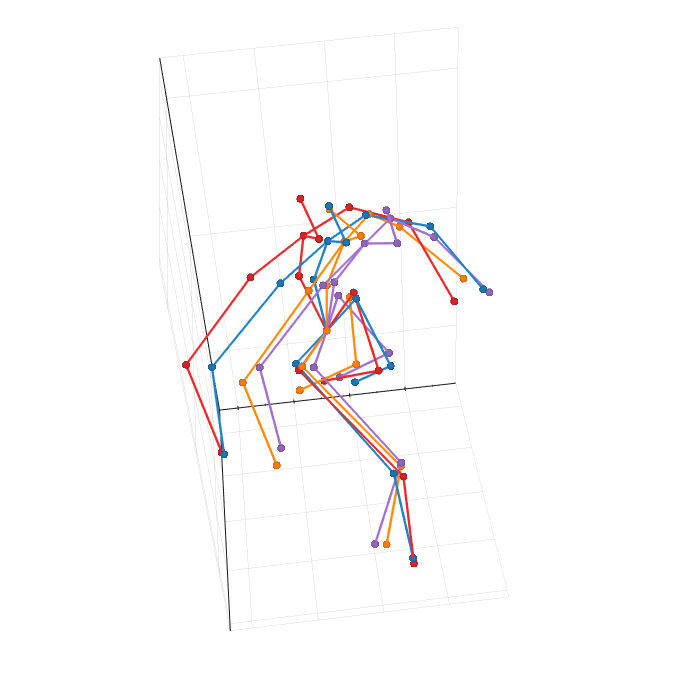}
         \includegraphics[width=\textwidth,keepaspectratio]{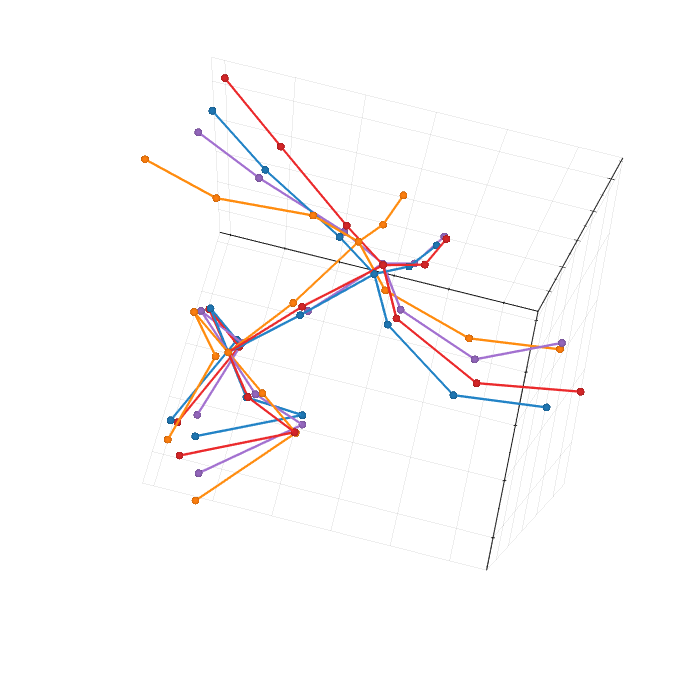}
         \includegraphics[width=\textwidth,keepaspectratio]{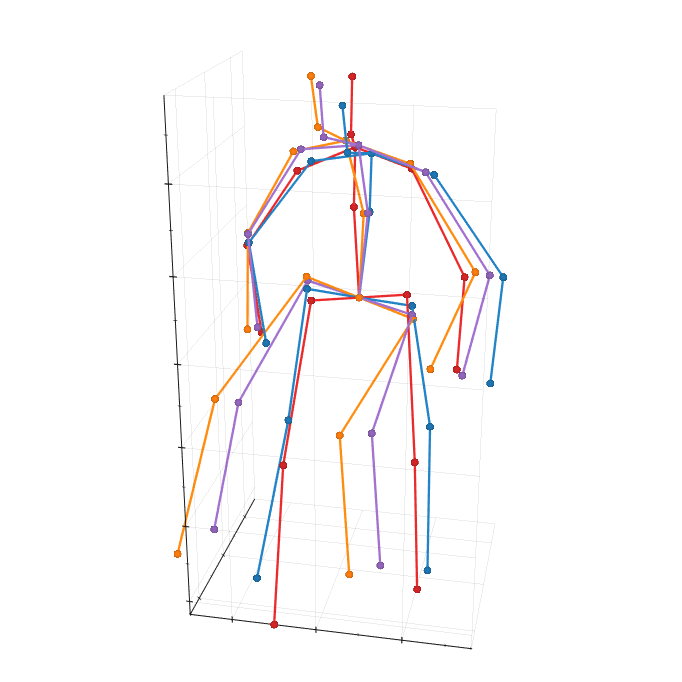}
         \includegraphics[width=\textwidth,keepaspectratio]{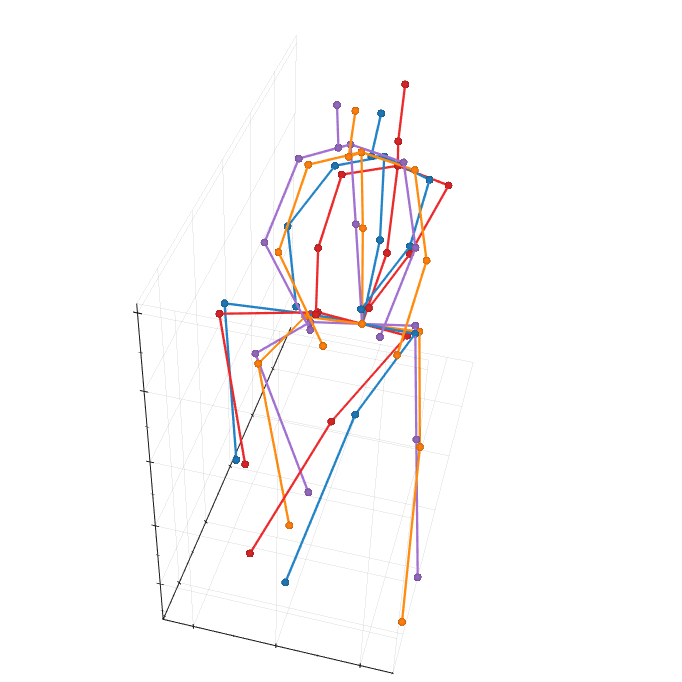}
    \caption*{3D View}
    \end{subfigure}
    
 	\caption{	\label{fig:final} Qualitative results from  3DPW (top 2), MPI-INF-3DHP (bottom 2) datasets. The last column shows our shows the pelvis centred 3D joint locations. Here, the ground truth joint locations are represented in red with PLIKS in blue, HybrIK in orange and CLIFF in purple.}
\end{figure*}

In Table~\ref{tab:main}, we provide quantitative results for previous 3D human shape and pose estimation results. We use the best results reported in all the other works for our comparison. 
Our network outperforms all previous state-of-the-art techniques on MPI-INF-3DHP and 3DPW datasets along with the lowest MPJPE on Human3.6M. Further fine-tuning with the AGORA dataset results in a significant performance improvement. We also provide results of our network when trained on a larger HRNet-W48~\cite{hrnet} backbone following previous works~\cite{cliff,pymafx2022,meshgraph}.
We provide qualitative results of our approach comparing prior methods~\cite{cliff,hybrik} in Fig.~\ref{fig:final}. Though all methods shown in Fig.~\ref{fig:final}  align well in the 2D projected space, it is apparent that the prior methods fail to align well in 3D space due to perspective warping. The effect of perspective warping is observed significantly when the human is off-centered. On the MPI-INF-3DHP dataset, the advantage of incorporating PLIKS is more pronounced since the images are captured with a wide Field-of-View (FOV)~($66^{\circ}{\sim}90^{\circ} $) showing a $26.3$~mm MPJPE improvement.

In Table~\ref{tab:agora}, we list quantitative results on the AGORA benchmark. It can be seen that the proposed approach significantly outperforms all prior methods. AGORA additionally measures the F1 score to get Normalized Mean Joint Error (NMJE) and
Normalized Mean Vertex Error (NMVE), which penalizes missed or faulty human detection. Here the human bounding box and camera parameters are not provided for the test set. We make use of YOLO-v5~\cite{yolo,h4w} for person detection. From Table~\ref{tab:agora}, PLIKS was fine-tuned only on AGORA, whereas, PLIKS$^\dagger$ was trained on all the 2D and 3D datasets which helps in generalizing to real-world images. As the focal length is not available during the test, we set $f{=}1$~m for all images. Making use of the estimated focal length from CamCalib~\cite{spec} on PLIKS$^\dagger$ shows a $3.4$~mm NMJE improvement (PLIKS$^\ddag$).

Most methods reported in Table~\ref{tab:main} generate root relative meshes using a weak perspective camera. As our results are in absolute camera coordinates,  we present root localization results in  Table~\ref{tab:root}. We report previous results on Human3.6M provided from ~\cite{rootnet} on the mean root position error (MRPE). Here baseline~\cite{mpiinf,lcr} depth is obtained using the least squares fit from the predicted 3D joints and its corresponding 2D projections. 
Though~\cite{rootnet} has explicitly been designed to learn the absolute depth, we observe comparable results along the depth while performing better in the horizontal and vertical directions. Using 3DPW and AGORA for training results in significant improvement of the root localization error. We also report the Absolute PCK result on the MuPoTs-3D~\cite{mupo} dataset in Table~\ref{tab:root2}. Note that the results are not strictly comparable due to differences in network backbones, the training datasets, and the joint prediction techniques.

\begin{table}[t]
\centering
\footnotesize
\begin{tabular}{l@{}|c|c@{}c|c@{}c}
\multicolumn{1}{c|@{}}{Method} & \multicolumn{1}{c|}{F1$\uparrow$} & \multicolumn{1}{c}{NMVE$\downarrow$} & \multicolumn{1}{@{}c|}{NMJE$\downarrow$} & \multicolumn{1}{c}{MVE$\downarrow$} & \multicolumn{1}{@{}c}{MPJPE$\downarrow$} \\ \hline
SPIN~\cite{spin}       & 0.77 & 193.4 & 199.2 & 148.9 & 153.4 \\\myrowcolour
SPEC~\cite{spec}      & 0.84 & 113.6 & 118.8 & 103.4 & 108.1 \\
BEV~\cite{bev}        & 0.93 & 108.3 & 113.2 & 100.7 & 105.3 \\ \myrowcolour 
H4W~\cite{h4w} & 0.94 & 90.2  & 95.5  & 84.8  & 89.8  \\ 
CLIFF~\cite{cliff} & 0.91 & 83.5  & 89.0  & 76.0  & 81.0  \\ \myrowcolour   \hline
PLIKS      & 0.94 & 76.8  & 81.5  & 72.2  & 76.6  \\ 
PLIKS$^\dagger$     & 0.94 & 78.3  & 83.0  & 73.6  & 78.0  \\ \myrowcolour
PLIKS$^\ddag$                      & 0.94                    & {74.9}            & {79.6}             & {70.4}           & {74.8}             \\
PLIKS$^\ddag$(HR48)                      & 0.94                    & \textbf{71.6}            & \textbf{76.1}             & \textbf{67.3}           & \textbf{71.5}             \\ \hline
\end{tabular}
\caption{\label{tab:agora} Reconstruction errors on the AGORA test set. All results are taken from the official evaluation platform. PLIKS$\dagger$  was trained with all 2D and 3D datasets.  PLIKS$\ddag$ represents the usage of CamCalib~\cite{spec} for focal length estimation during evaluation.   
}
\end{table}

\begin{table}[b]
\footnotesize

\centering
\begin{tabular}{l|cc|cc}
     & \multicolumn{2}{c|}{3DPW-OCC}         & \multicolumn{2}{c}{3DOH}              \\ \cline{2-5} 
     & \multicolumn{1}{c|}{MPJPE$\downarrow$} & PA-MPJPE$\downarrow$ & \multicolumn{1}{c|}{MPJPE$\downarrow$} & PA-MPJPE$\downarrow$ \\ \hline
DOH~\cite{3doh}  & \multicolumn{1}{c|}{-}      & 72.2     & \multicolumn{1}{c|}{-}     & 58.5     \\\myrowcolour
EFT~\cite{eft}  & \multicolumn{1}{c|}{94.4}  & 60.9     & \multicolumn{1}{c|}{75.2}  & 53.1     \\ 
PARE~\cite{pare} & \multicolumn{1}{c|}{90.5}  & 56.6     & \multicolumn{1}{c|}{63.3}  & 44.3     \\ \myrowcolour \hline
PLIKS(R50) & \multicolumn{1}{c|}{\textbf{86.1}}  & \textbf{53.2}     & \multicolumn{1}{c|}{\textbf{51.5}}  & \textbf{39.3}    \\ \hline
\end{tabular}\caption{\label{tab:occ}Evaluation on occlusion datasets 3DPW-OCC and 3DOH.}
\end{table}

\begin{table*}[ht]

\centering
\footnotesize
\begin{tabular}{lc|cc|cc|cc}
\multicolumn{2}{c|}{} & \multicolumn{2}{c|}{3DPW} & \multicolumn{2}{c|}{Human3.6M} & \multicolumn{2}{c}{MPI-INF-3DHP} \\ \cline{3-8} 
\multicolumn{1}{c|}{Module} & Camera                   & MPJPE$\downarrow$ & PA-MPJPE$\downarrow$ & MPJPE$\downarrow$ & PA-MPJPE$\downarrow$ & MPJPE$\downarrow$ & PA-MPJPE$\downarrow$ \\ \hline
\multicolumn{1}{l|}{ARE}    & (wP)                    & 83.6  & 50.5     & 53.6  & 39.4     & 94.6  & 61.4     \\
\multicolumn{1}{l|}{ARE(R50)}    & (wP)                    & 87.1  &   53.2   & 54.5  & 40.5     & 95.9  & 61.7     \\ \hline
\multicolumn{1}{l|}{PLIKS}  & (wP)                    & 85.5  & 50.6     & 51.3  & 36.2     & 93.4  & 63.2     \\ \hline
\multicolumn{1}{l|}{PLIKS}  & $f{=}{\sqrt{(W^2+H^2)}}$ (P)& 81.8  & 51.7     & 47.8 & 34.8    & 83.2  & 61.4     \\ 
\multicolumn{1}{l|}{PLIKS}  & Known (P) & 81.8  & 51.9     & 48.9 & 34.8    & 76.9 & 60.6    \\\hline
\end{tabular}\caption{\label{tab:abalation}Ablation studies by varying network setting. Here, (wP) and (P) refers to the weak-perspective and perspective camera model respectively.} 
\end{table*}

\subsection{Occlusion Analysis}
To validate the stability under occlusion, we evaluate PLIKS on the object-occluded benchmark dataset 3DOH50K~\cite{3doh} and the person-occluded dataset 3DPW-OCC~\cite{3dpw}. Following~\cite{pare}, we use only COCO, Human3.6M, and 3DOH datasets for training. To arrive at a fair comparison, we also train a network with ResNet50~\cite{r50} as the backbone. We observe better occlusion performance on both backbones from Table~\ref{tab:occ} for our approach. This can be attributed to the fact, that our approach is intrinsically trained on dense correspondences.

\subsection{Ablation Studies}
In this context, we evaluate the individual components of our approach. For all experiments, we use only the COCO, Human3.6M, and MPI-INF-3DHP datasets for training and evaluation on all 3D datasets. We use the 3DPW validation set to determine the best model. In Table~\ref{tab:abalation}, we provide all the results for our ablation studies. The ARE module on its own can reconstruct 3D models provided the shape and camera parameters are regressed by the network. Hence, we evaluate its performance without the PLIKS module. Here, we observe that it performs poorly on all metrics for all datasets except for the PA-MPJPE of 3DPW. Next, on the same ARE module, we replace the HRNet-32 backbone with a ResNet-50 backbone. In this setting, similar to the occlusion test,
HRNet performs better than ResNet. Next, we evaluate the importance of accounting for camera intrinsics by training a network with the PLIKS module but with all cropped images given as input to the network having a fixed focal length of $f{=}0.3~\mathrm{m}$. The low focal length value violates the assumption of a weak-perspective setting. We observe slight improvements compared to that of  ARE on the Human3.6M, whereas we see on-par results with the other datasets.  Finally, we determine the effect of using a focal length of $f{=}\sqrt{W^2+H^2}$ proposed in~\cite{wpc} only during inference. While there is no significant difference in 3DPW and Human3.6M, we observe a drop in performance for the MPI-INF-3DHP dataset compared to PLIKS with known camera intrinsics. This is due to the actual FOV being larger than the estimated FOV. Our ablations clearly demonstrate the importance of incorporating camera intrinsics into a network.

\section{Conclusion}
In this paper, we bridge the gap between 2D correspondence and body mesh estimation by creating a pipeline from which the inverse kinematics can be solved in closed form. This can be effectively leveraged to use the full perspective projection rather than having to rely on weak-perspective counterparts. Our approach yields considerable improvements in both root-relative and absolute-3D estimation for human pose estimation. PLIKS further enables our method to be fully differentiable facilitating end-to-end training. We validated the effectiveness of our method on various 3D pose and shape datasets and achieved state-of-the-art on multiple benchmarks. Due to the inherent nature of a built-in solver, we can extend our work with additional constraints like using multi-view systems for even higher reconstruction accuracy or temporal constraints for smooth video reconstruction should be feasible and foster future research in this direction.

\paragraph*{Disclaimer}
The concepts and information presented in this article are based on research and are not commercially available.

{\small
\bibliographystyle{ieee_fullname}
\bibliography{egbib}
}
\clearpage

\section*{Appendix}

\maketitle

\section{Introduction}
In this material, we provide implementation details and analysis of focal lengths and regularizers for our method. We further discuss the benefits of using a solver for human pose estimation utilizing constraints. Additionally, we present more qualitative results, to show the performance of PLIKS and to explore its failure scenarios.

\subsection{Datasets}\label{datasets}
\noindent\textbf{COCO:} COCO~\cite{coco} is a large-scale in-the-wild 2D key-point dataset. We use this for training. We make use of pseudo-ground truth SMPL annotations provided by EFT~\cite{eft}.
\\\noindent\textbf{MPI-INF-3DHP:} MPI-INF-3DHP is an indoor multi-view and outdoor scene dataset for 3D human pose estimation. We make use of SMPL multi-view fits by SPIN~\cite{spin}. We use this for training and evaluation. 
\\\noindent\textbf{Human3.6M:} Human3.6M~\cite{h36m}
is an indoor, multi-view 3D human pose
estimation dataset.  We follow the standard practice~\cite{hmr,spin} where subjects S1, S5, S6, S7, and S8 are used
for training while S9 and S11 are the test subjects. We follow Protocol 2 using only the front-facing cameras.\\\noindent\textbf{3DPW:} 3DPW~\cite{3dpw} is a challenging outdoor benchmark for 3D pose and shape estimation. To get a fair comparison with previous state-of-the-art~\cite{pare,hybrik}, we use 3DPW training data for 3DPW experiments. We make use of a subset of this dataset 3DPW-OCC following~\cite{pare} for the occlusion benchmark.
\\\noindent\textbf{AGORA:} AGORA~\cite{agora}  is a synthetic dataset with accurate SMPL models fitted to 3D scans. The test set is not publicly available, here the evaluation is performed on the official platform. For both training and testing, we use the images of resolution $1280\times720$.
\\\noindent\textbf{3DOH:} 3DOH~\cite{3doh} is an object-occluded dataset. We use this to train and evaluate only for occlusion benchmark.
\\\noindent\textbf{MuPoTs-3D:} MuPoTs-3D~\cite{mupo} is a mixed indoor and outdoor multi-person dataset consisting of 20 sequences showing people performing various actions and interactions. We use this for evaluating the absolute root error.

\begin{figure}
\begin{center}
\includegraphics[width=1\linewidth]{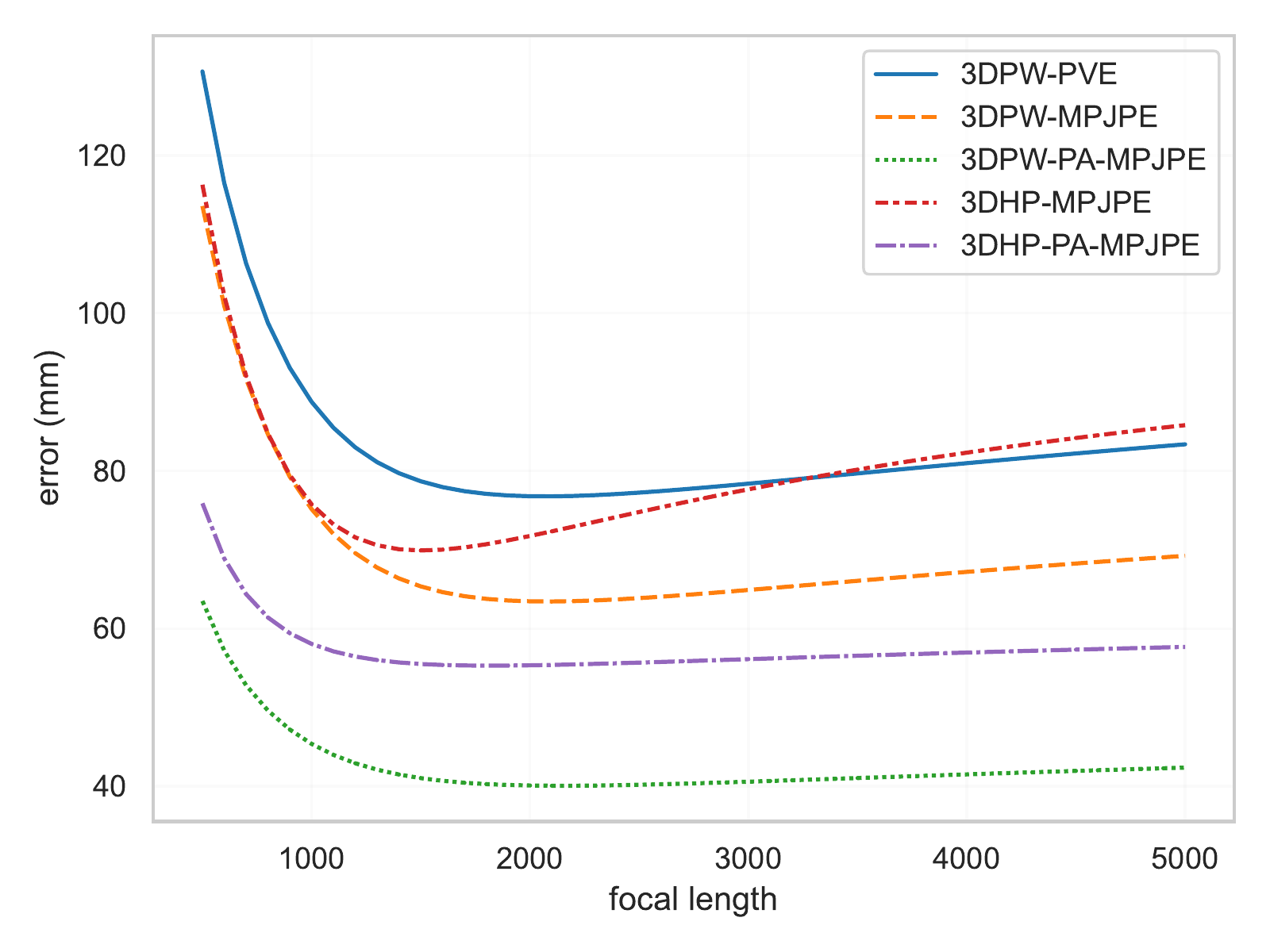}
\end{center}
   \caption{Impact on focal length on estimation errors when using the 3DPW~\cite{3dpw} and MPI-INF-3DHP~\cite{mpiinf} dataset. 
}
\label{fig:focal_vary}
\end{figure}

\subsection{Network Training}
As the entire pipeline is differentiable, the network is trained end-to-end. We split the training into two steps, pre-training (ARE) and training (PLIKS) to accelerate the network training speed. In pre-training, we train exclusively with the ARE module, and optimize only with respect to the mesh and network predicted parameters ($\vecb{\Tilde{\beta}}, \vecb{\Tilde{\theta}}_{k}$) by minimizing,
\begin{equation}\label{eq_obj}
   L = \omega_1 L_{\theta} + \omega_2 L_{\beta} + \omega_3L_{2d} + \omega_4L_{3d} + \omega_5L_{M}.
\end{equation}
Following previous work~\cite{hmr,spin,i2l}, we employ standard mesh losses to supervise the training process. Here, $L_{\theta}$ is the $\mathrm{L}2$ loss between the predicted pose and ground truth (GT) pose. Similarly, $L_{\beta}$ is the $\mathrm{L}2$ loss between the predicted shape and GT shape. $L_{2d}$, $L_{3d}$ and $L_{M}$ are the $\mathrm{L}1$ loss between predictions and GT 2D joint re-projection, 3D joints and, the mesh vertex in image space respectively. To supervise the 2D annotations, the 
predicted 3D joints are projected by the weak-perspective camera  $\vecb{\Tilde{c}}$ as predicted by the network. 

During training we make use of the PLIKS module. Due to the presence of the linear solver in PLIKS, we observe numerical instability in the early stages of training, i.e. the pixel-aligned vertex predictions are not adequately consistent for the solver, making the reconstruction ill-posed. To keep the error within bounds, we add strong shape and pose regularizers for two epochs. In this stabilization period, the shape regularizer $\omega_\beta$  exponentially decays from 1 to 0.1. We further add a pose-constraint to the objective function of PLIKS (Eq.~\eqref{eq_argmin}), such that $\omega_\theta\sum{|\mathbf{{\Delta R}}_k|} \approx \mathbf{I}$. As a consequence, the additional rotation $\mathbf{{\Delta R}}_k$ obtained during the stabilization period is constrained to be close to zero. Similar to $\omega_\beta$, we decay $\omega_\theta$ from 1 to 0. For training, we use the same objective function from  Eq.~\eqref{eq_obj} to minimize the mesh and the analytically predicted parameters ($\vecb{{\beta}}, \vecb{{\theta}}_{k}$).

\begin{equation}\label{eq_argmin}
    \underset{ \mathbf{{\Delta R}}_k,\vecb{\beta},\vecb{t}_k}{\mathrm{argmin}}|| \vecb{w}^k\Big(\vecb{i}^k - \mathbf{\hat{K}} ( \mathbf{{\Delta R}}_k {\vecb{x}_r^k}   + \vecb{\beta} {\mathbf{B}_r^k}  +  \vecb{t}_k\mathbf{W}_r^k)\Big)||_2 {+} \omega_\beta||\vecb{\beta}||_2.
\end{equation}

\subsection{Implementation Details}
PyTorch~\cite{pytorch} is used for implementation. For all our experiments we initialize the HRNet~\cite{hrnet} backbone with weights pre-trained on the MPII~\cite{mpii} dataset, which exhibits faster convergence during training. We use the Adam optimizer~\cite{adam} with a mini-batch size of 32. The learning rate at pre-training is set to $1e^{-4}$, whereas, while training the entire pipeline it is initialized to $5e^{-5}$. The network is pre-trained for 20 epochs, stabilized for 2 epochs, and then finally trained for further 30 epochs. We set the learning rate to $1e^{-5}$ while fine-tuning with the 3DPW~\cite{3dpw} or AGORA~\cite{agora} dataset. For fine-tuning, we use the previous pre-trained network as the starting point. This is to accelerate convergence and correct the 3D inaccuracies from the pseudo-GT labels.
It takes around 3-5 days to train on a single NVIDIA Tesla V-100-16GB GPU. We set $\omega_1$, $\omega_2$, $\omega_3$, $\omega_4$, and $\omega_5$ to 1, 0.05, 4, 8, and 4, respectively. As the pseudo-GT labels from EFT~\cite{eft}, are defined with respect to weak-perspective projection, we reduce $\omega_1$, $\omega_2$, $\omega_4$, and $\omega_5$ by a factor of 0.1 for the 2D dataset.

\section{Ablations}
Here we discuss the effects of shape regularizer and effects of focal length estimation.

\subsection{Regularizer}
\begin{figure}[!b]
\begin{center}
\includegraphics[width=1\linewidth]{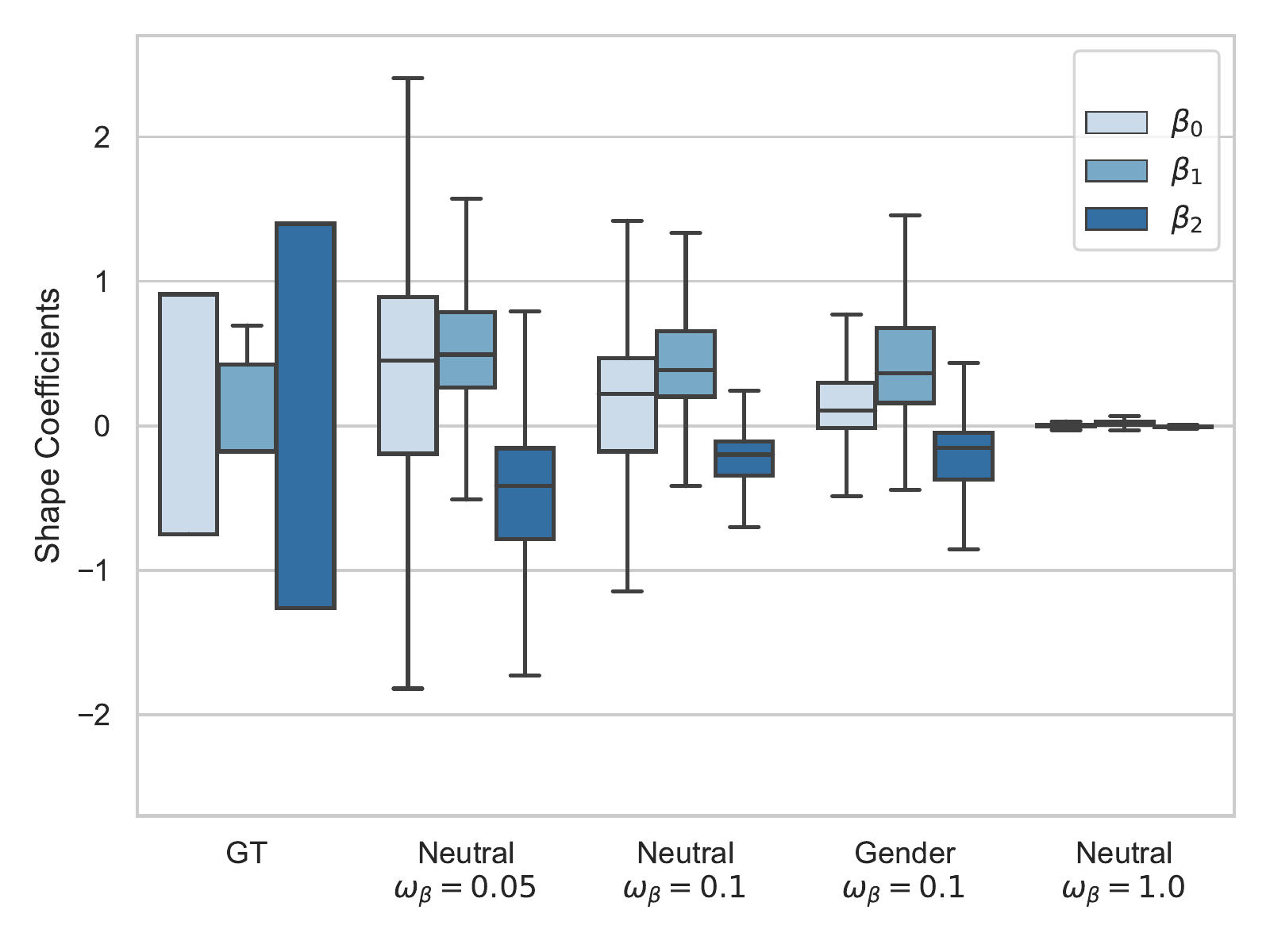}
\end{center}
   \caption{Effect of the regularizer weight used during training on the shape coefficients. Picking a higher $\omega_\beta$ reduces the error, but causes the network to output meshes getting progressively closer to the identity representation. Here neutral represents the neutral SMPL model, and gender refers to the gender-specific model on the 3DPW~\cite{3dpw} dataset. 
}
\label{fig:beta_vary}
\end{figure}

To demonstrate the importance of a strong regularizer, we perform a similar experiment (from Sec 4.1) where we add random noise to the GT of the mesh vertices from the 3DPW~\cite{3dpw} test set. Here we vary the shape regularizer weights $\omega_\beta$ and observe the final MPJPE obtained. From Table~\ref{tab:reg}, it is evident that larger weights for $\omega_\beta$ is more robust to noise. However, training the network using larger weights has its own drawbacks as shown in Figure~\ref{fig:beta_vary}. The network forces the shape components $\vecb{\beta}$ to always be close to zero. 
As the shape $\vecb{\beta}$ is determined by a solver, it enables us to switch to a male, female, or neutral model seamlessly by replacing the shape coefficients $\mathbf{B}$ during inference. For our training, we set $\omega_\beta=0.1$, as this is a good mixture between stability and shape variations.

\subsection{Focal Length}
We conduct experiments on the 3DPW and MPI-INF-3DHP test sets by varying the focal lengths. As shown in Fig.~\ref{fig:focal_vary}, PLIKS is robust to a wide range of focal lengths when the FOV is small (e.g., 3DPW), but it suffers from the effects of perspective warping on large focal lenghts for wide FOV images (e.g., MPI-INF-3DHP). Using CamCalib~\cite{spec} on the MPI-INF-3DHP to determine the FOV and consequently the focal length of the image, we could only obtain a reduction in MPJPE of $72.01$~mm, i.e., a drop of just 3\%.
In particular, when, the ground truth camera matrix is known, our approach can be expected to yield optimal performance.

\begin{figure*}[ht]
	\centering
    \includegraphics[width=1\linewidth]{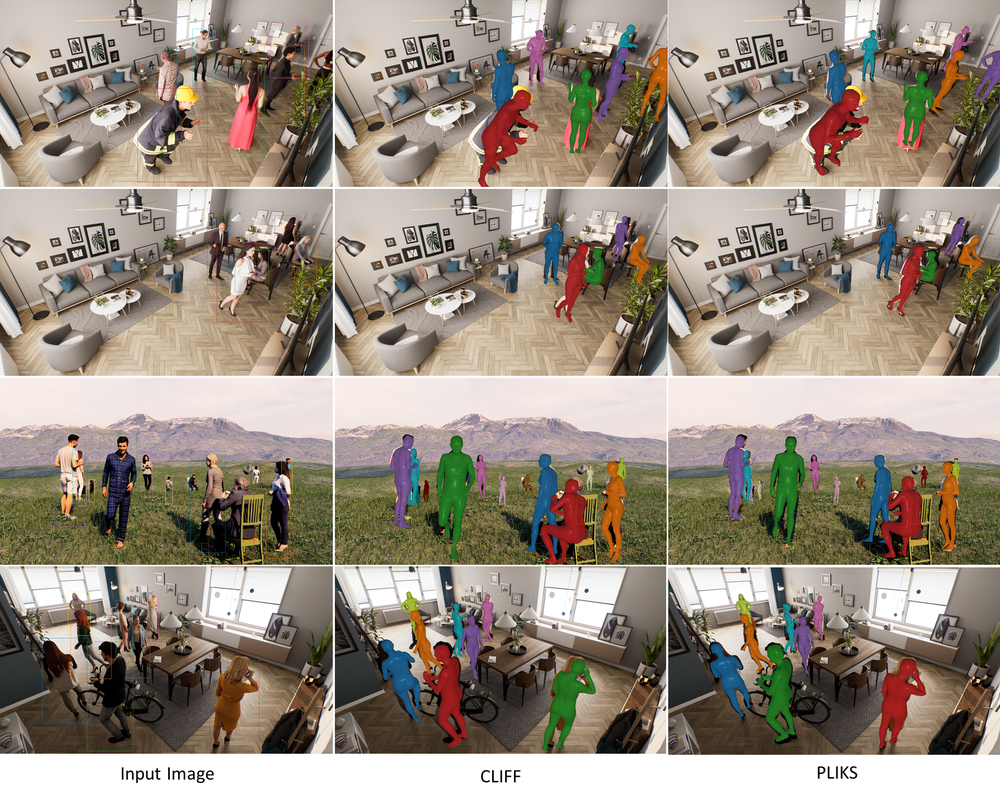}

 	\caption{	\label{fig:agora} Qualitative results from  AGORA test set.}
\end{figure*}

\begin{table}[]
\centering
\begin{tabular}{l|c|c|c}
\hline
       & $\pm10~\mathrm{mm}$ & $\pm20~\mathrm{mm}$ & $\pm30~\mathrm{mm}$ \\ \hline
$\omega_\beta=2.0$   & 17.3   & 22.6   & 34.9   \\\myrowcolour
$\omega_\beta=1.0$    & 18.4   & 37.3   & 64.0   \\ 
$\omega_\beta=0.1$  & 122    & 254    & 322.1  \\ \hline
\end{tabular}\caption{\label{tab:reg}Ground truth errors in the presence of per vertex noise ranging from $\pm10 mm$ to $\pm30 mm$ and the effect of using a shape regularizer, $\omega_{\beta}$.  }
\end{table}

\section{Qualitative Results}

In this section, we show comparisons to SOTA
methods on AGORA and provide more qualitative results on various other datasets.

\subsection{Qualitative Comparison}
We display several examples of PLIKS on the AGORA test set in Fig.~\ref{fig:agora}. We use YOLO~\cite{yolo} for the bounding box estimation and CamCalib~\cite{spec} for the focal length estimation. The images demonstrate that PLIKS performs better than previous approaches, by aligning the bodies well in 3D as well as 2D.  

\subsection{Inference Modification}
As mentioned in the main paper, one of the strengths of our method is the application of constraints during inference.
Here, we discuss a proof-of-concept for two use cases, where we show the benefits of using a solver without any retraining of the network. We discuss dynamic shape and translation constraints.

\begin{figure*}
\begin{center}
\includegraphics[width=1\linewidth]{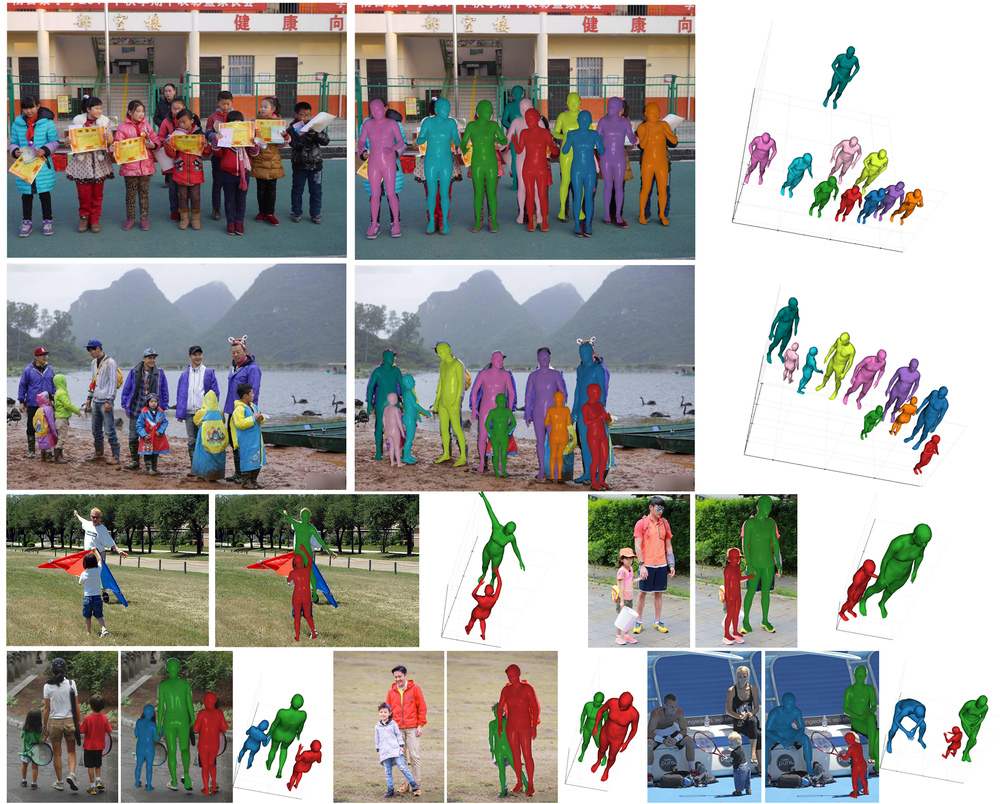}
\end{center}
   \caption{Example images with dynamic shape during inference. Set of input images, overlay and, 3D view.}
\label{fig:fig_kid1}
\end{figure*}

\paragraph{Dynamic Shape}
Although our network was trained only on a neutral SMPL model with 10 shape components, it can make use of other shape models during inference if they follow the same design principle as SMPL. As an extreme scenario, we show the application using the kid-SMPL model~\cite{agora, bev}. The kid-SMPL is an extended version of the SMPL model supporting children by linearly blending the SMPL and  Skinned Multi-Infant Linear Model (SMIL)~\cite{siml} by a weighting factor $\alpha \in (0,1)$~\cite{agora}. Here, larger weights represents infants, while smaller weights are associated with adults. For simplicity, we denote the kid-SMPL model as having 11 shape components.

Qualitative results of using the kid-SMPL model on the Relative Human (RH) dataset~\cite{bev} are shown in Fig.~\ref{fig:fig_kid1}. The only modification performed was adapting the shape coefficients $\mathbf{B}_r^k$ in Eq.~\ref{eq_argmin} from the SMPL to their kid-SMPL counterparts. In that context, we further empirically set $\omega_\beta$ to 0.5. From the RH dataset we employ the GT age classifier, i.e., we use SMPL for adults, and kid-SMPL for child or infant. We observe visually satisfactory results, with sufficiently reliable depth reasoning. A top-down approach~\cite{bev} or a simple age classifier could be designed to determine the age as a future work.

\begin{figure*}
\begin{center}
\includegraphics[width=0.8\linewidth]{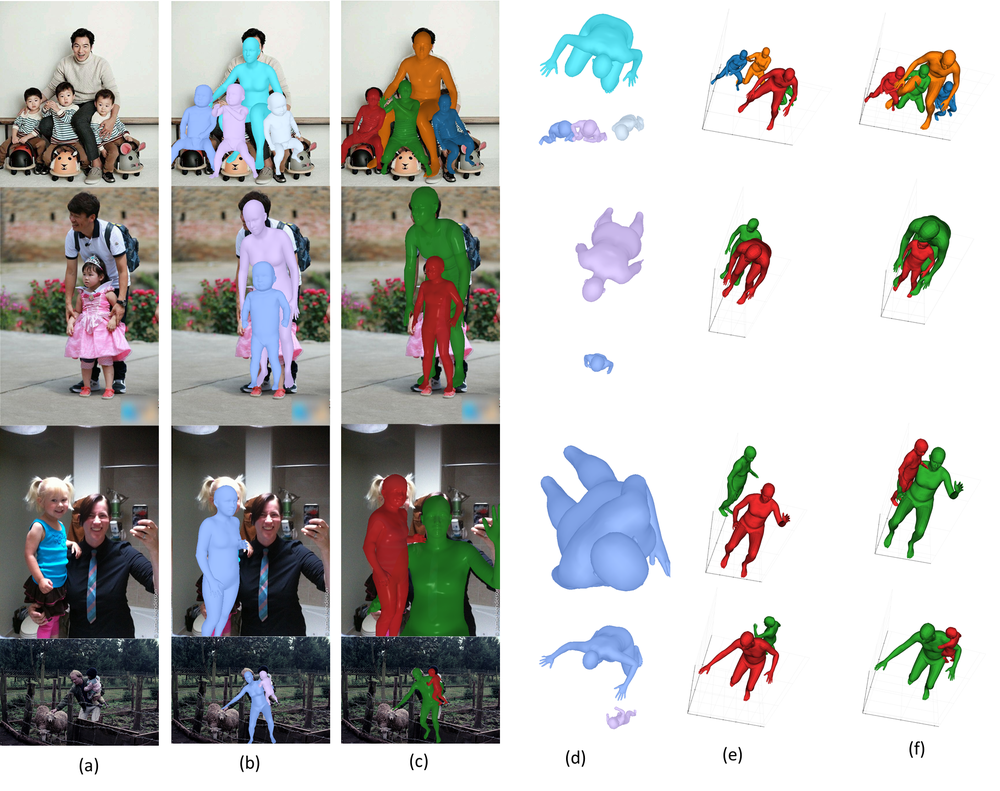}
\end{center}
   \caption{Example images with translation constraints during inference. (a) Input Image, (b,c) 3D overlay from BEV~\cite{bev} and PLIKS respectively, (d) 3D view of the model from BEV~\cite{bev}, (e,f) 3D view of the model from PLIKS without and with using the translation constraint.}
   \vspace{-20pt}
\label{fig:fig_kid2}
\end{figure*}

\paragraph{Translation Constraints}
Previous examples of just using dynamic shapes is not a complete solution, due to the ill-posed nature of the problem. This is quite evident from the fifth column of Fig.~\ref{fig:fig_kid2}. As a proof-of-concept, we show the application of translation constraints during inference. We add a simple depth constraint to Eq.~\ref{eq_argmin} as $\omega_t \vecb{t}_{0,z}^{k} = \omega_t \vecb{t}_{0,z}^{a}$. Here, $\vecb{t}_{0,z}^{a}$ is the root depth of the adult in the image, and  $\vecb{t}_{0,z}^k$ is the constrained setting for the root-depth of the kids in the image, with $\omega_t$ being a weighting factor. We make the assumption that the children in the images are standing close to the adults. The solver optimizes the shape such that the translation constraint is satisfied. We empirically set $\omega_t$ to 0.2. Though, strictly not comparable, we visualize the results of BEV~\cite{bev} in Fig.~\ref{fig:fig_kid2}. There, all images are in fact from the RH training set on which BEV was trained. We quickly add that this is not a real-world solution to the problem, but it emphasizes the importance of using constraints during inference or training. As future work, one could make use of the RH dataset with the depth-level information by adding a top-down approach~\cite{bev} for better constraints.

\subsection{Failure Mode}
\begin{figure*}[hb]
\begin{center}
\includegraphics[width=0.85\linewidth]{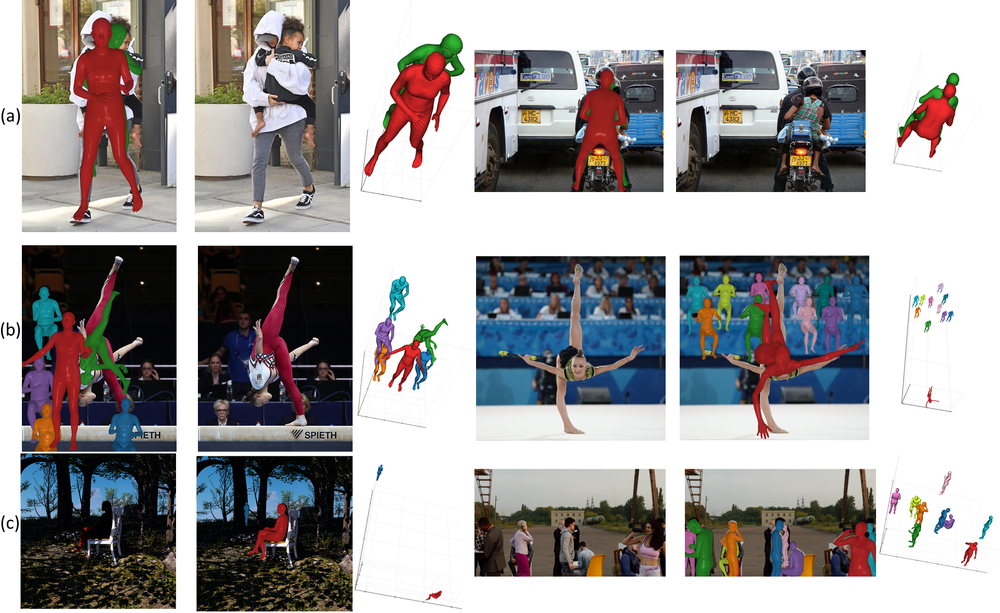}
\end{center}
   \caption{Example of failure cases.}
   \vspace{-20pt}
\label{fig:fail}
\end{figure*}

In Fig.~\ref{fig:fail}, we show a few examples where
PLIKS fails to reconstruct reasonable human body poses. The failure cases range from (a) too many people in the crop, (b) extreme poses not seen in training, and (c) extreme occlusion.

\begin{figure*}
\begin{center}
\includegraphics[width=1\linewidth]{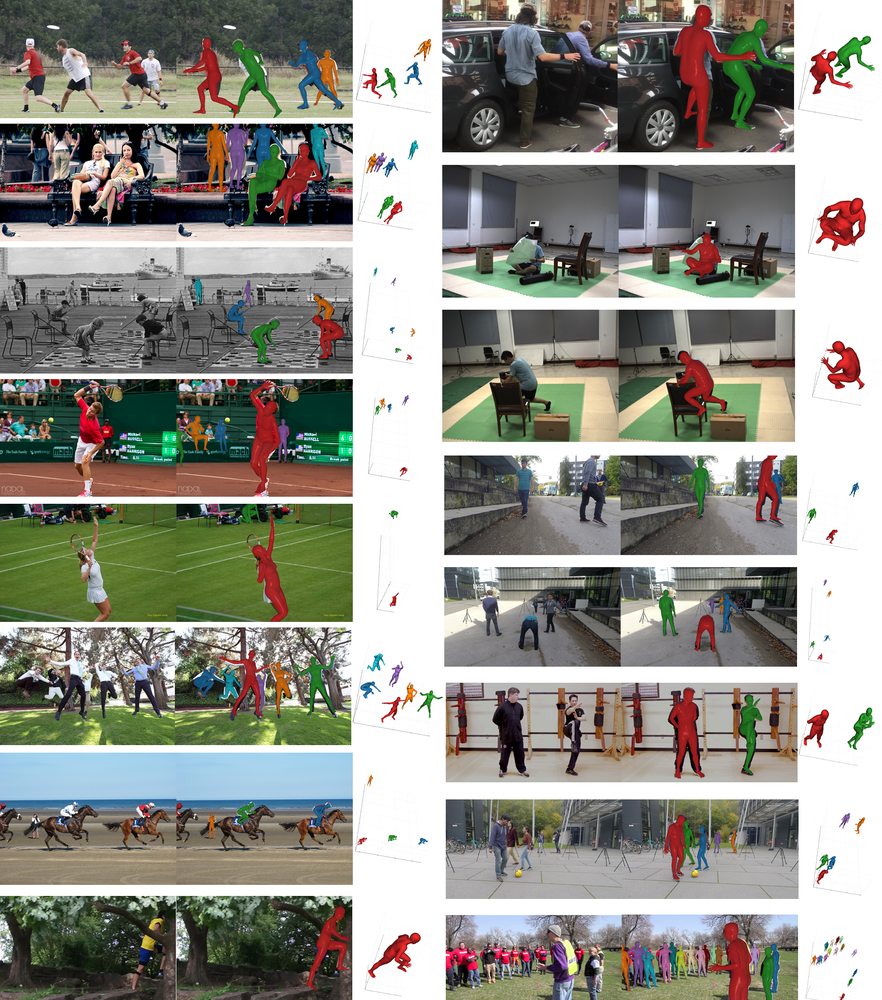}
\end{center}
   \caption{Additional qualitative results of PLIKS from COCO~\cite{coco}, MPII~\cite{mpii}, 3DPW~\cite{3dpw}, 3DOH~\cite{3doh} and MuPoTs-3D~\cite{mupo}. Set of challenging input images, overlay and, 3D view.}
\label{fig:fig_qual}
\end{figure*}
\clearpage

\end{document}